\documentclass[10pt,twocolumn,letterpaper]{article}
\usepackage{cvpr}
\usepackage[dvipsnames]{xcolor}

\definecolor{cvprblue}{rgb}{0.21,0.49,0.74}
\usepackage[pagebackref,breaklinks,colorlinks,citecolor=cvprblue]{hyperref}

\usepackage{arydshln}
\usepackage{multirow}
\usepackage{amsmath} 
\usepackage{amsfonts} 
\usepackage{pifont}
\usepackage{bbm}
\usepackage{xcolor}
\usepackage{soul}
\usepackage[accsupp]{axessibility} 
\usepackage{times}

\definecolor{best}{RGB}{255,218,218} 
\definecolor{second}{RGB}{217,218,253} 
\newcommand{\hlcolor}[2][best]{\sethlcolor{#1}\hl{#2}}

\def\ie{\textit{i.e.}}
\def\eg{\textit{e.g.}}

\newcommand{\task}{low-level }
\newcommand{\name}{{\textit{Diff-Plugin }}}

\newlength\savewidth\newcommand\shline{\noalign{\global\savewidth\arrayrulewidth
		\global\arrayrulewidth 1pt}\hline\noalign{\global\arrayrulewidth\savewidth}}
\newcommand{\tablestyle}[2]{\setlength{\tabcolsep}{#1}\renewcommand{\arraystretch}{#2}\centering\small}

\definecolor{demphcolor1}{gray}{.6}
\newcommand{\demphs}[1]{\textcolor{demphcolor1}{#1}}

\newcommand{\yu}[1]{\textcolor[rgb]{0.0, 0.0, 0.0}{#1}}
\newcommand{\ryn}{\textcolor[rgb]{0,0,0}}

\newcommand{\ke}{\textcolor[rgb]{0,0,0}}
\newcommand{\fang}[1]{\textcolor[rgb]{.0,.0,.0}{#1}}

\newcommand{\figref}[1]{Fig.~\ref{#1}}

\newcommand{\secref}[1]{Sec.~\ref{#1}}
\newcommand{\tableref}[1]{Table~\ref{#1}}

\title{Diff-Plugin: Revitalizing Details for Diffusion-based Low-level Tasks}

\author{Yuhao Liu$^{1}$, \ \ Zhanghan Ke$^{1, \dagger}$, \ \ Fang Liu$^{1,\dagger}$, \ \ Nanxuan Zhao$^{2}$, \ \ Rynson W.H. Lau$^{1, \dagger}$ \\
$^{1}$City University of Hong Kong, \ \ $^{2}$Adobe Research\\
}

\begin{document}

\twocolumn[{
\maketitle
\begin{center}
    \captionsetup{type=figure}
    \setlength{\tabcolsep}{2pt}
    \footnotesize
    \includegraphics[width=1\linewidth]{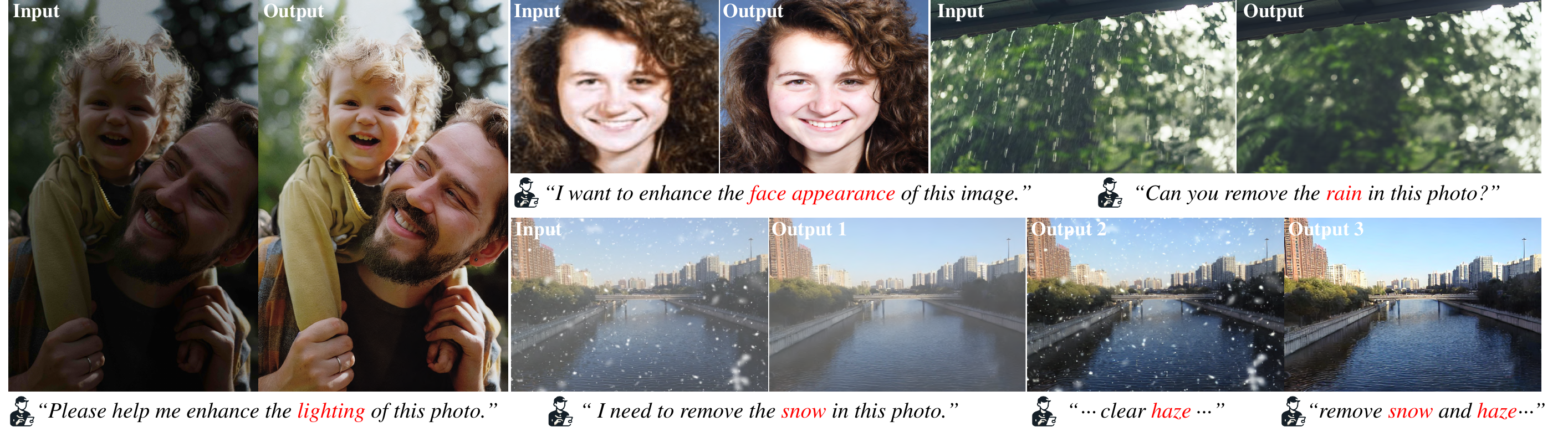}
    \vspace{-6mm}
    \captionof{figure}{Real-world applications of \name visualized across distinct single-type and one multi-type \task vision tasks. \name allows users to selectively conduct interested  \task vision tasks via natural languages and can generate high-fidelity results.
    }
    \label{fig:teaser}
    \vspace{2mm}
\end{center}
}]

\makeatletter{\renewcommand*{\@makefnmark}{}
\footnotetext{$^{\dagger}$Joint corresponding authors. 
This project is in part supported by a GRF grant (Grant No.: 11205620) from the Research Grants Council of Hong Kong.
\makeatother}

\begin{abstract}

Diffusion models trained on large-scale datasets have achieved remarkable progress in image synthesis. However, due to the randomness in the diffusion process, they often struggle with handling diverse low-level tasks that require details preservation. 
To overcome this limitation, we present a new \name framework to enable a single pre-trained diffusion model to generate high-fidelity results across a variety of low-level tasks. 
Specifically, we first propose a lightweight Task-Plugin module with a dual branch design to 
\ke{provide task-specific priors, guiding the diffusion process in preserving image content.}
We then propose a Plugin-Selector that can automatically select different Task-Plugins based on the text instruction, allowing users to edit images by indicating multiple low-level tasks with natural language.
We conduct extensive experiments on 8 \task vision tasks. The results demonstrate the superiority of Diff-Plugin over existing methods, particularly in real-world scenarios. 
Our ablations further validate that Diff-Plugin is stable, schedulable, and supports robust training across different dataset sizes. 
Project page: \href{https://yuhaoliu7456.github.io/Diff-Plugin}{https://yuhaoliu7456.github.io/Diff-Plugin}

\end{abstract}

\section{Introduction}
\label{sec:intro}

Over the past two years, diffusion models \cite{song2020denoising,ho2020denoising,dhariwal2021diffusion,ho2022classifier} have achieved unprecedented success in image generation and shown potential to become vision foundation models. 
Recently, many works~\cite{li2023your,karazija2023diffusion,zhao2023generative,zhang2023inversion,parmar2023zero,kawar2023imagic,brooks2023instructpix2pix} have demonstrated that diffusion models trained on large-scale text-to-image datasets can already understand various visual attributes and provide versatile visual representations for downstream tasks, \eg, image classification~\cite{li2023your}, segmentation~\cite{karazija2023diffusion,zhao2023generative}, translation~\cite{zhang2023inversion,parmar2023zero}, and editing~\cite{kawar2023imagic,brooks2023instructpix2pix}.

However, due to the inherent randomness in the diffusion process, existing diffusion models cannot maintain consistent contents to the input image and thus fail in handling \task vision tasks.  
To this end, some methods~\cite{tumanyan2023plug,parmar2023zero} propose to utilize input images as a prior via the DDIM Inversion~\cite{song2020denoising} strategy when editing images, but they are unstable when the scenes are complex. 
Other methods~\cite{saharia2022palette,ren2023multiscale,wang2023dr2,yi2023diff,guo2023shadowdiffusion} attempt to train new diffusion models on task-specific datasets from scratch, limiting them to solve only a single task.

In this work, we observe that an accurate text prompt describing the goal of the task can already instruct a pre-trained diffusion model to address many \task tasks, but typically leads to obvious content distortion, as illustrated in \figref{fig:observation}. 
Our insight to this problem is that task-specific priors containing both guidance information of the task and spatial information of the input image can adequately guide pre-trained diffusion models to handle \task tasks while maintaining high-fidelity content consistency. 
To harness this potential, we propose \textit{Diff-Plugin}, the first framework enabling a pre-trained diffusion model, such as stable diffusion~\cite{rombach2022high}, to accommodate a variety of \task  tasks without compromising its original generative capability. 

\textit{Diff-Plugin} consists of two main components. First, it includes a lightweight Task-Plugin module to help extract task-specific priors. 
The Task-Plugin is bifurcated into the Task-Prompt Branch (TPB) and the Spatial Complement Branch (SCB). While TPB distills the task guidance prior, orienting the diffusion model towards the specified vision task and minimizing its reliance on complex textual descriptions, SCB leverages task-specific visual guidance from TPB to assist the spatial details capture and complement, enhancing the fidelity of the generated content. 
Second, to facilitate the use of multiple different Task-Plugins, \textit{Diff-Plugin} includes a  Plugin-Selector to allow users to choose their desired Task-Plugins through text inputs (visual illustrations are depicted in \figref{fig:teaser}). 
To train the Plugin-Selector, we employ multi-task contrastive learning \cite{radford2021learning}, using task-specific visual guidance as pseudo-labels. This enables the Plugin-Selector to align different visual embeddings with task-specific text inputs, thereby bolstering the robustness and user-friendliness of the Plugin-Selector.

To thoroughly evaluate our method, we conducted extensive experiments on eight diverse \task vision tasks. Our results affirm that \name is not only stable across different tasks but also exhibits remarkable schedulability, facilitating text-driven multi-task applications. 
Additionally, \name showcases its scalability, adapting to various tasks across datasets of varying sizes, from less than 500 to over 50,000 samples, without affecting existing trained plugins. 
Finally, our results also show that the proposed framework outperforms existing diffusion-based methods both visually and quantitatively, and achieves competitive performances compared to regression-based methods.

\begin{figure}[t!]
    \centering
    \includegraphics[width=1\linewidth]{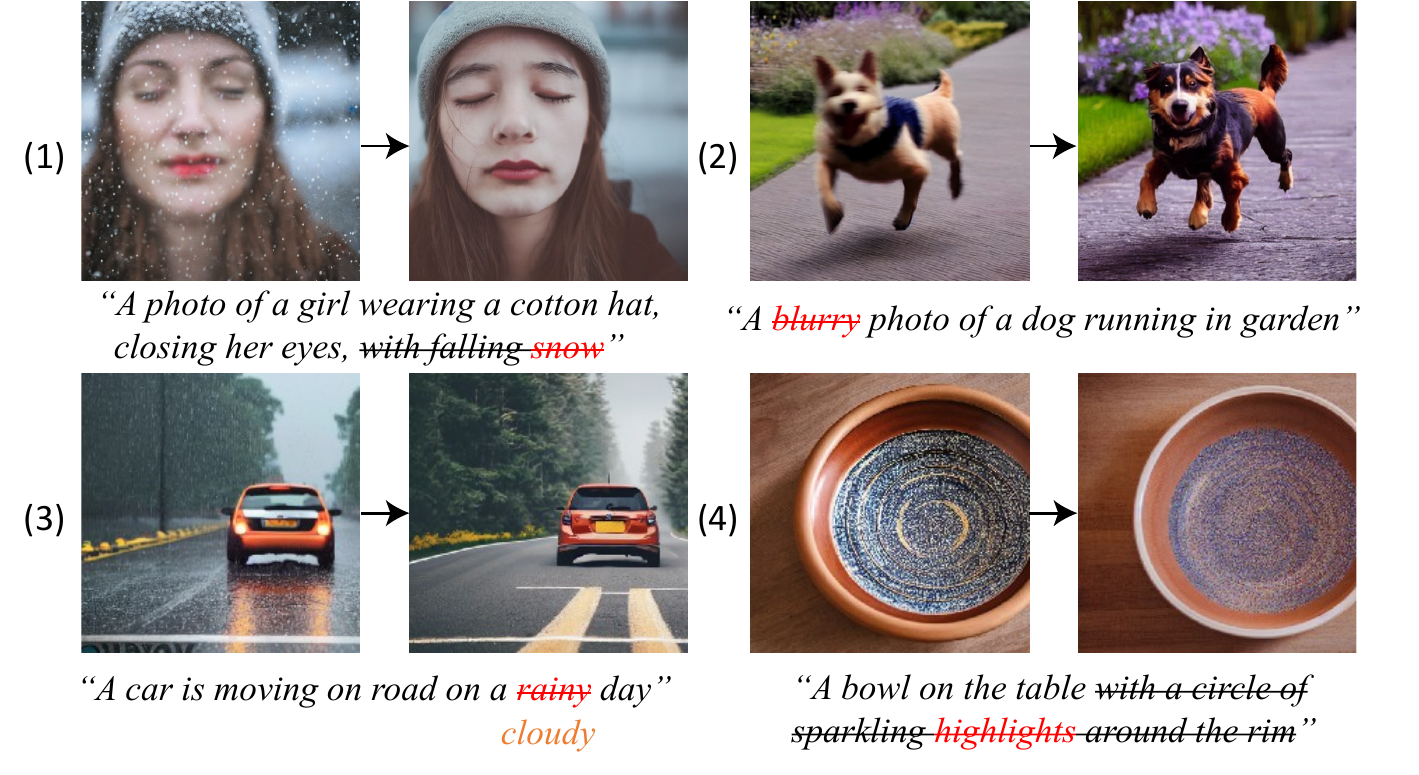}
    \vspace{-6mm}
    \caption{ Stable Diffusion (SD) \cite{rombach2022high} results on four low-level vision tasks: desnowing, deblurring, deraining, and highlight removal. Each sub-figure illustrates a two-step process: 
    First, we generate the left image using SD with a full-text description, where task-critical attributes are highlighted in {\color{red}{red}}. 
    Then, we remove unwanted attributes (indicated with \st{strikethrough}), optionally add new attributes (denoted with {\color{orange}{orange}} word), and employ the \textit{img2img} function in SD, using the left image as a condition to generate the edited image on the right. 
    We observe that while SD can grasp rich attributes of various low-level tasks and create content consistent with descriptions, its inherent randomness often leads to content change in further editing. 
    For instance, in sub-fig (1), besides addressing the primary task-related degradation (\eg, snow), SD also alters unrelated content (\eg, face profile).
    }
    \label{fig:observation}
\end{figure}

Our key contributions are summarized as follows: 
\begin{itemize}
    \item We present \textit{Diff-Plugin}, the first framework to enable a pre-trained diffusion model to perform various \task tasks while maintaining the original generative abilities. 
    \item We propose a Task-Plugin, a lightweight dual-branch module designed for injecting task-specific priors into the diffusion process, to enhance the fidelity of the results. 
    \item We propose a Plugin-Selector to select the appropriate Task-Plugin based on the text provided by the user.  This extends to a new application that can allow users to edit images via text instructions for \task vision tasks.
    \item We conduct extensive experiments on eight tasks, demonstrating the competitive performances of \textit{Diff-Plugin} over existing diffusion and regression-based methods.
\end{itemize}

\section{Related Works}
\label{sec:related}

\textbf{Diffusion models} \cite{sohl2015deep,song2019generative} have been applied to image synthesis \cite{ho2020denoising,song2020denoising,dhariwal2021diffusion,ho2022classifier} and achieved remarkable success. 
With extensive text-image data \cite{schuhmann2022laion} and large-scale language models \cite{raffel2020exploring,radford2021learning}, diffusion-based text-guided image synthesis \cite{nichol2021glide,ramesh2022hierarchical,saharia2022photorealistic,rombach2022high,balaji2022ediffi} has become even more compelling. 
Leveraging the text-guided synthesis diffusion model, several approaches harness the generative prowess for text-driven editing. Zero-shot approaches \cite{hertz2022prompt,parmar2023zero,tumanyan2023plug} rely on a correct initial noise \cite{song2020denoising} and manipulate the attention map to edit specified content at precise locations. 
Tuning-based strategies strive to balance between image fidelity and generated diversity through optimized DDIM inversion \cite{wallace2023edict}, attention tuning \cite{kumari2023multi}, text-image coupling \cite{zhang2023sine,ruiz2023dreambooth,kawar2023imagic} and prompt tuning \cite{mokady2023null,dong2023prompt,gal2022image}. 
Conversely, InstructP2P \cite{brooks2023instructpix2pix, zhang2023magicbrush} generates paired data through latent diffusion \cite{rombach2022high} and prompt-to-prompt \cite{hertz2022prompt} for training and editing. 
However, the randomness in the diffusion process and the absence of task-specific priors render them infeasible for \task vision tasks that require details preservation.

\noindent \textbf{Conditional generative models} 
use various external inputs to ensure output consistency with the conditions. Training-free methods \cite{couairon2023zero,xie2023boxdiff} can generate new contents at specified positions by manipulating attention layers, yet with limited condition types. 
Fine-tuning-based approaches inject additional guidance to the pre-trained diffusion models by training a new diffusion branch \cite{zhang2023adding,mou2023t2i,zhao2023uni} or the whole model \cite{avrahami2023spatext}. 
Despite the global structural consistency, these methods cannot ensure high-fidelity between output and input image details due to the randomness and generative nature.

\noindent \textbf{Diffusion-based \task methods} can be grouped into zero-shot and training-based. The former can borrow generative priors from pre-trained denoising diffusion-based generative models \cite{ho2020denoising} to solve linear \cite{kawar2022denoising,wang2022zero} and/or non-linear \cite{fei2023generative,chung2023diffusion} \fang{image restoration} tasks, but often produce poor results on real-world data. 
The latter usually train or fine-tune an individual model for different tasks via task-dependent designs, such as super-resolution \cite{saharia2022image, xia2023diffir}, 
JPEG compression \cite{saharia2022palette}, 
deblurring \cite{whang2022deblurring,ren2023multiscale}, 
face restoration \cite{wang2023dr2,zhao2023towards}, 
low-light enhancement \cite{yi2023diff,zhang2023unified, jiang2023low}, and shadow removal \cite{guo2023shadowdiffusion}. 
Concurrent works, StableSR \cite{wang2023exploiting} and DiffBIR \cite{lin2023diffbir}, use a learnable conditional diffusion branch with degraded or restored images to train diffusion models specifically for blind face restoration. 
In contrast, our framework enables one pre-trained diffusion model to handle a variety of \task tasks by equipping it with lightweight task-specific plugins.

\noindent \textbf{Multi-task models}  can learn complementary information across different tasks, \eg, object detection and segmentation \cite{he2017mask}, rain detection and removal \cite{yang2017deep}, adverse weather restoration \cite{zhu2023learning,ye2023adverse, ozdenizci2023restoring} and blind image restoration \cite{potlapalli2023promptir,li2022all}. 
However, these methods can only handle the pre-defined tasks after training.  Instead, our \textit{Diff-Plugin} is flexible and can integrate new tasks through task-specific plugins, as our Task-Plugins are trained individually. Hence, when adding new \task tasks to \textit{Diff-Plugin}, we only need to add the pre-trained Task-Plugins to the framework, without the need to retrain the existing ones.

\section{Methodologies}
\label{sec:method}

In this section, we first review the diffusion model formulations (\secref{subsec:pre}). Then, we introduce our \textit{Diff-Plugin} framework (\secref{subsec:diff_plugin}), which developed from our newly proposed Task-Plugin (\secref{subsec:plugin}) and Plugin-Selector (\secref{subsec:scheduler}).

\subsection{Preliminaries}
\label{subsec:pre}

The diffusion model consists of a forward process and a reverse process. 
In the forward process, given a clean input image $\mathbf{x}_{0}$, the diffusion model progressively adds Gaussian noise to it to get noisy image $\mathbf{x}_t$ at time-step $t\in \{0,1,...,T \}$, as $\mathbf{x}_t = \sqrt{\bar{\alpha}_t} \mathbf{x}_0+\sqrt{1-\bar{\alpha}_t} \boldsymbol{\epsilon}_{t}$, where $\bar{\alpha}_t$ is the pre-defined scheduling variable and $ \boldsymbol{\epsilon}_t \sim \mathcal{N}(\mathbf{0}, \mathbf{I})$ is the added noise. 
In the reverse process, the diffusion model performs iteratively remove noise from a standard Gaussian noise $\mathbf{x}_T$, and finally estimating a clean image $\mathbf{x}_0$. 
This is typically employed to train a noise prediction network $\boldsymbol{\epsilon}_\theta$, with supervision informed by the noise $ \boldsymbol{\epsilon}_t$, 
as $\mathcal{L}=\mathbb{E}_{\mathbf{x}_0 \text{,} t \text{,} \epsilon \sim \mathcal{N}(0\text{,}1)} \left[ \left\|\boldsymbol{\epsilon}-\boldsymbol{\epsilon}_\theta\left(\mathbf{x}_t \text{,} \ t\right)\right\|_2^2\right]$.

\begin{figure}[t!]
    \centering
    \includegraphics[width=1.0\linewidth]{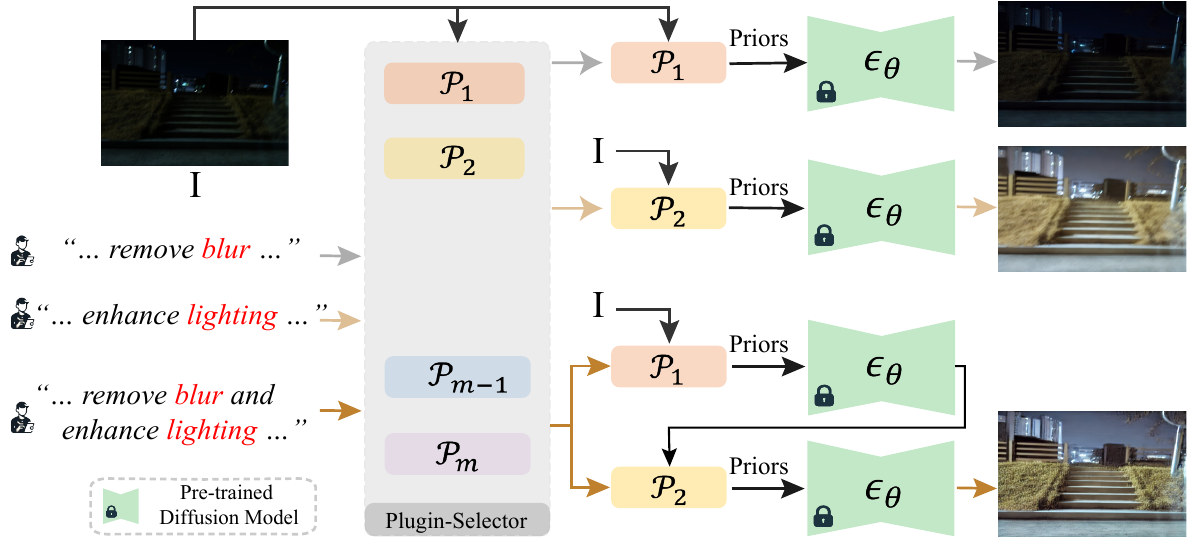}
    \vspace{-5mm}
    \caption{Schematic illustration of the \textit{Diff-Plugin} framework. 
    \textit{Diff-Plugin} identifies appropriate Task-Plugin $\mathcal{P}$ based on the user prompts, extracts task-specific priors, and then injects them into the pre-trained diffusion model to generate the user-desired results.}
    \vspace{-2mm}
    \label{fig:pipeline}
\end{figure}

\subsection{Diff-Plugin}
\label{subsec:diff_plugin}

Our key observation is the inherent zero-shot capability of pre-trained diffusion models in performing \task vision tasks, enabling them to generate diverse visual content without explicit task-specific training. 
However, this capability faces limitations in more nuanced task-specific editing. 
For example, in the desnowing task, while the model should ideally only remove snow and leave other contents unchanged, as shown in \figref{fig:observation}, the inherent randomness of the diffusion process often leads to unintended alterations in the scene beyond just snow removal. This inconsistency arises from the model's lack of task-specific priors, which are crucial for precise detail preservation in \task vision tasks.

Inspired by modular extensions in NLP \cite{xu2023small,xiao2023plug} and GPT-4 \cite{gpt_plugins}, which utilize plug-and-play tools to enhance the capabilities of large language models for downstream tasks without compromising their core competencies, 
we introduce a novel framework, \textit{Diff-Plugin}, \ryn{based on a similar idea}. This framework integrates several lightweight plugin modules, termed Task-Plugin, into the pre-trained diffusion models for various low-level tasks. \ryn{Task-Plugins} are crafted to provide essential task-specific priors, guiding the models to produce high-fidelity and task-consistent content.
In addition, while diffusion models can generate content based on text instructions for targeted scenarios, they lack the ability to schedule Task-Plugins for different low-level tasks. Even existing conditional generation methods \cite{zhang2023adding,qin2023unicontrol} can only specify different generation tasks through input conditional images. Thus, to facilitate smooth text-driven task scheduling and enable the switching between different Task-Plugins for complex workflows, \textit{Diff-Plugin} \ryn{includes a Plugin-Selector to allow} users to choose and schedule appropriate Task-Plugins with textual commands.

\figref{fig:pipeline} depicts the \textit{Diff-Plugin} framework. 
Given an image, users specify the task through a text prompt, either singular or multiple, and the Plugin-Selector identifies the appropriate Task-Plugin for it. 
The Task-Plugin then processes the image to extract the task-specific priors, guiding the pre-trained diffusion model to produce user-desired outcomes. 
For more intricate tasks beyond the scope of a single plugin, \textit{Diff-Plugin} breaks them down into sub-tasks with \ryn{a predefined mapping table}. 
Each sub-task is tackled by a designated Task-Plugin, showcasing the framework's capability to handle diverse and complex \ryn{user requirements}.

\subsection{Task-Plugin}
\label{subsec:plugin}

As illustrated in \figref{fig:plugin}, our Task-Plugin module is composed of two branches: a Task-Prompt Branch (TPB) and a Spatial Complement Branch (SCB). 
The TPB is crucial for providing task-specific guidance to the pre-trained diffusion model, akin to using text prompts in text-conditional image synthesis \cite{rombach2022high}. 
We employ visual prompts, extracted via the pre-trained CLIP vision encoder \cite{radford2021learning}, to direct the model's focus towards task-relevant \yu{patterns (\eg, rain streaks for deraining and snow flakes for desnowing)}. 
Specifically, for an input image $\mathbf{I}$, the encoder $\textit{Enc}_{I}(\cdot)$ first extracts general visual features, which are then distilled by the TPB to yield discriminative visual guidance priors $\mathbf{F}^{p}$:
\begin{equation}
    \mathbf{F}^{p} = \textit{TPB}(\textit{Enc}_{I}(\mathbf{I})) \text{,}
\label{eq:tpb}
\end{equation}
where TPB, comprising three MLP layers with Layer Normalization and LeakyReLU activations (except for the final layer), ensures the retention of only the most task-specific attributes. This approach aligns $\mathbf{F}^{p}$ with the textual features the diffusion model typically uses in its text-driven generation process, thus facilitating better task alignment \yu{for Plugin-Selector}. Furthermore, using visual prompts simplifies the user's role by eliminating the need for complex text prompt engineering, which is often challenging for specific vision tasks and sensitive to minor textual variations \cite{xu2023prompt}.

However, the task-specific visual guidance prior $\mathbf{F}^{p}$, while \yu{crucial for prompting global semantic attributes, is not sufficient for preserving fine-grained details.}
In this context, DDIM Inversion plays a pivotal role by providing initial noise that contains information about the image content. 
Without this step, the inference would rely on random noise devoid of image content, resulting in less controllable results in the diffusion process. 
\yu{However, the inversion process is unstable and time-consuming.} \yu{To alleviate this}, we introduce the SCB  to extract and enhance spatial details preservation effectively. 
\fang{We} utilizes the pre-trained VAE encoder \cite{esser2021taming} $\textit{Enc}_{V}(\cdot)$, to capture full content of input image $\mathbf{I}$, \fang{denoted} as $\mathbf{F}$. 
This comprehensive image detail, when combined with the semantic guidance from $\mathbf{F}^{p}$, is then processed by our SCB \yu{to distill the spatial feature  $\mathbf{F}^{s}$:}
\begin{equation}
\mathbf{F}^{s} = \textit{SCB}(\mathbf{F}\text{,} \ \mathbf{F}^{t}\text{,} \ \mathbf{F}^{p})=\textit{Att}(\textit{Res}(\mathbf{F}\text{,} \ \mathbf{F}^{t})\text{,} \ \mathbf{F}^{t}\text{,} \ \mathbf{F}^{p}) \text{,}
\label{eq:SCB}
\end{equation}
where $\mathbf{F}^t$ is time embedding used to denote the varied time step in diffusion process. The $\textit{Res}$ and $\textit{Att}$ blocks represent the standard ResNet and Cross-Attention transformer blocks, from the diffusion model \cite{rombach2022high}. The output from $\textit{Res}$ is utilized as the Query features and $\mathbf{F}^{p}$ acts as both Key and Value features in the cross-attention \fang{layer}.

We then introduce the task-specific \yu{visual guidance} prior $\mathbf{F}^{p}$ into the cross-attention layers of the diffusion model, where it serves to \yu{direct the model's generation process toward the specific requirements of the low-level vision task.}
Following this, we directly incorporate the distilled spatial prior $\mathbf{F}^{s}$  into the final stage of the decoder as a residual. 
This placement is based on our experimental observations in \tableref{tab:plugin_position}, which indicated that the fidelity of spatial details in the stable diffusion \cite{rombach2022high} tends to decrease from the shallow layers to the deeper ones. 
By adding $\mathbf{F}^{s}$ at this specific stage, we effectively counteract this tendency, thereby enhancing the preservation of fine-grained spatial details. 

To train the Task-Plugin modules, we adopt the denoising loss as defined in \cite{rombach2022high}, introducing the task-specific priors into the diffusion denoising training process: 
\begin{equation}
    \mathcal{L}=\mathbb{E}_{\boldsymbol{z}_0\text{,} t \text{,} \mathbf{F}^{p} \text{,} \mathbf{F}^{s} \text{,}  \epsilon \sim \mathcal{N}(0\text{,}1)}\left[\| \epsilon-\epsilon_\theta\left(\boldsymbol{z}_t\text{,} \ t \text{,} \ \mathbf{F}^{p} \text{,} \ \mathbf{F}^{s}\right) \|_2^2\right] \text{,}
\label{eq:denois_loss}
\end{equation}
where $\mathbf{z}_t = \sqrt{\bar{\alpha}_t} \mathbf{z}_0+\sqrt{1-\bar{\alpha}_t} \boldsymbol{\epsilon}_{t}$ represents the noised version of the latent-space image at time $t$, and $\mathbf{z}_0$, the latent-space representation of the ground truth image $\hat{\mathbf{I}}$, is obtained as $\mathbf{z}_0=\textit{Enc}_{V}(\hat{\mathbf{I}})$. This loss function ensures that the Task-Plugin is effectively trained to incorporate the task-specific priors in guiding the diffusion process.

\begin{figure}[t!]
    \centering
    \includegraphics[width=1\linewidth]{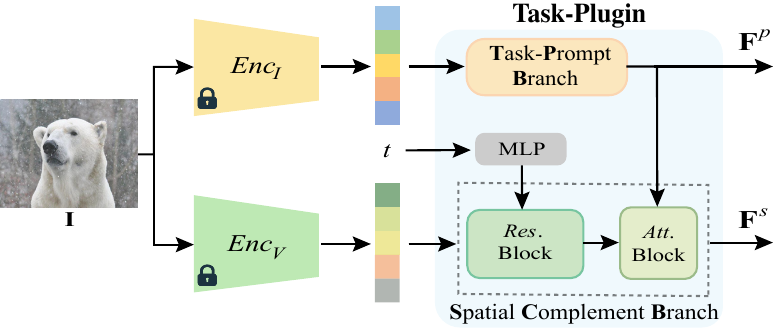}
    \vspace{-7mm}
    \caption{Schematic illustration of task-specific priors extraction via the proposed lightweight Task-Plugin. Task-Plugin processes three inputs: time step $t$, visual prompt from $\textit{Enc}_I(\cdot)$, and image content from $\textit{Enc}_V(\cdot)$. 
    It distills visual guidance $\mathbf{F}^p$ via a task-prompt branch and extracts spatial features $\mathbf{F}^s$ through a spatial complement branch, jointly for task-specific priors.
    }
    \label{fig:plugin}
\end{figure}

\subsection{Plugin-Selector}
\label{subsec:scheduler}

We propose the Plugin-Selector, enabling users to select the desired Task-Plugin using text input. 
For an input image $\mathbf{I}$ and a text prompt $\mathbf{T}$, we define the set of Task-Plugins as $\mathcal{P}=\{\mathcal{P}_1, \mathcal{P}_2, \cdots, \mathcal{P}_m\}$, 
with each $\mathcal{P}_i$ corresponding to a specific vision task, transforming $\mathbf{I}$ into task-specific priors $(\mathbf{F}^{p}_i, \mathbf{F}^{s}_i)$. 
Then, visual guidance $\mathbf{F}^{p}_i$ of each Task-Plugin is then cast to a new textual-visual aligned multi-modality latent space via a shared visual projection head $\textit{VP}(\cdot)$ and denoted as $\mathcal{V} = \{\boldsymbol{v}_1, \boldsymbol{v}_2, \cdots, \boldsymbol{v}_m\}$. 
Concurrently, $\mathbf{T}$ is encoded into a text embedding by $\textit{Enc}_{T}(\cdot)$  \cite{radford2021learning} and then projected to $\boldsymbol{q}$ using a textual project head $\textit{TP}(\cdot)$, aligning the textual and visual embedding. The process is formulated as:
\begin{equation}
    \boldsymbol{v}_i = \textit{VP}(\mathbf{F}_{i}^{p})\text{;} \quad \boldsymbol{q} = \textit{TP}(\textit{Enc}_{T}(\mathbf{T})) \text{.}
\end{equation}

We then compare the textual embedding $\boldsymbol{q}$ with each visual embedding $\boldsymbol{v}_i \in \mathcal{V}$ using cosine similarity function such that $\boldsymbol{s}_i = \text{sim}(\boldsymbol{v}_i, \boldsymbol{q})$,  yielding a set of similarity scores $\mathcal{S} = \{\boldsymbol{s}_1, \boldsymbol{s}_2, \cdots, \boldsymbol{s}_m\}$.
We select the Task-Plugin $\mathcal{P}_{\text{selected}}$ that meet a specified similarity threshold, $\theta$: 
 \begin{equation}
\mathcal{P}_{\text{selected}} = \{\mathcal{P}_i \mid \boldsymbol{s}_i \geq \theta \text{,}  \ \mathcal{P}_i \in \mathcal{P}\}.
\end{equation}

We adopt the $\mathbf{F}^{p}_i$ as the pseudo label and pair it with task-specific text to construct training data. We employ contrastive loss \cite{chen2020simple,radford2021learning} to optimize the vision and text projection heads, enhancing their capability to handle multi-task scenarios. 
This involves minimizing the distance between the anchor image and positive texts while increasing the distance from negative texts. 
For each image $\mathbf{I}$, a positive text relevant to its task (\eg, ``\textit{I want to remove rain}'' for deraining task) and  $N$ negative texts from other tasks (\eg, ``\textit{enhance the face}'' for face restoration) are sampled. 
The loss function for a positive pair of example $(i,j)$ is as follows:
\begin{equation}
    \ell_{i, j}=-\log \frac{\exp \left(\operatorname{sim}\left(\boldsymbol{v}_{i}\text{,}  \  \boldsymbol{q}_{j}\right) / \tau\right)}{\sum_{k=1}^{N+1} \mathbbm{1}_{[k_{c} \neq i_{c}]} \exp \left(\operatorname{sim}\left(\boldsymbol{v}_{i}\text{,}  \  \boldsymbol{q}_k\right) / \tau\right)} \text{,} 
\label{eq:scheduler}
\end{equation}
where $c$ represents the task type for each sample and $\mathbbm{1}_{[k_{c} \neq i_{c}]} \in  \{0 \text{,} 1 \}$ is an indicator function evaluating to 1 iff $k_{c} \neq i_{c}$. 
$\tau$ denotes a temperature parameter.

\begin{figure*}[htp]
\begin{center}
\begin{tabular}{c@{\hspace{1.7mm}}c@{\hspace{0.7mm}}c@{\hspace{0.7mm}}c@{\hspace{0.7mm}}c@{\hspace{0.7mm}}c@{\hspace{0.7mm}}c@{\hspace{0.7mm}}c@{\hspace{0.7mm}}c}
\multirow{1}{*}[1.1cm]{\rotatebox{90}{\fontsize{8pt}{\baselineskip}\selectfont Desnowing}}
&\includegraphics[width=0.116\linewidth,height=0.085\linewidth]{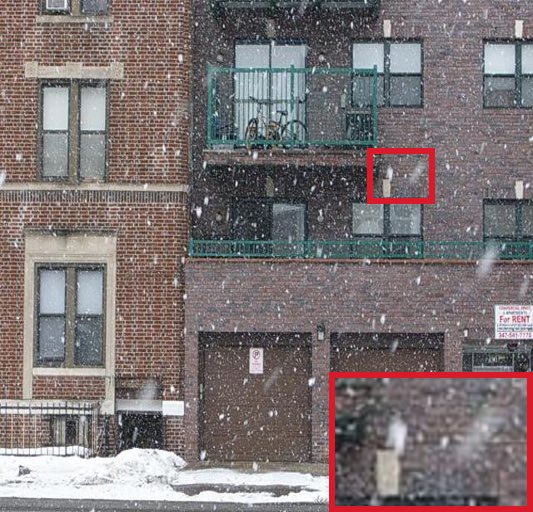}&
\includegraphics[width=0.116\linewidth,height=0.085\linewidth]{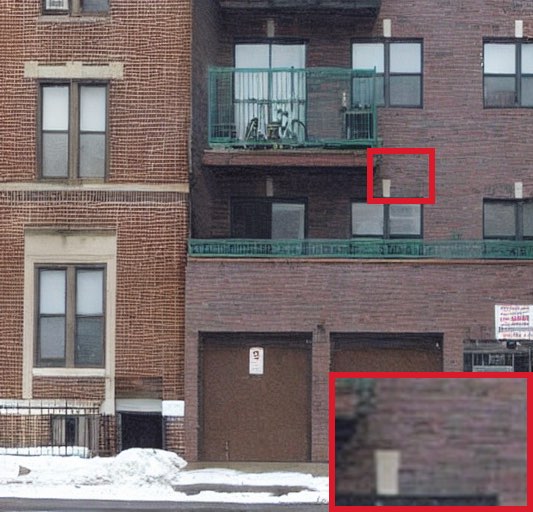}&
\includegraphics[width=0.116\linewidth,height=0.085\linewidth]{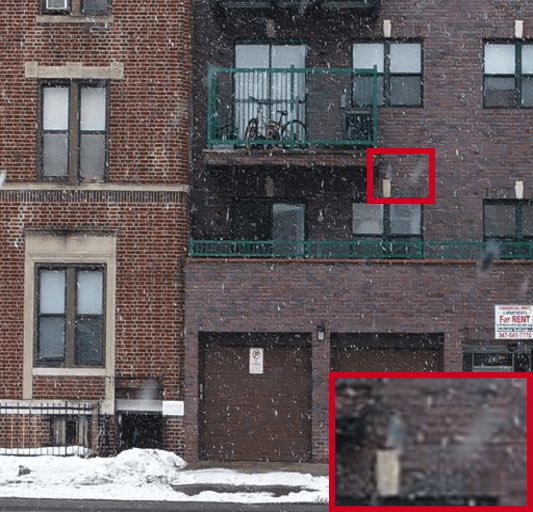}&
\includegraphics[width=0.116\linewidth,height=0.085\linewidth]{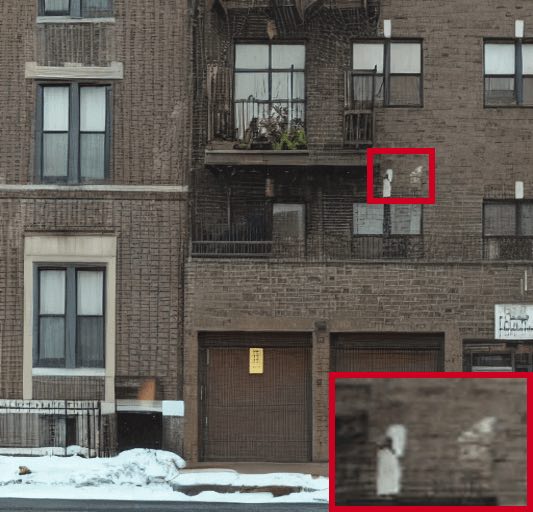}&
\includegraphics[width=0.116\linewidth,height=0.085\linewidth]{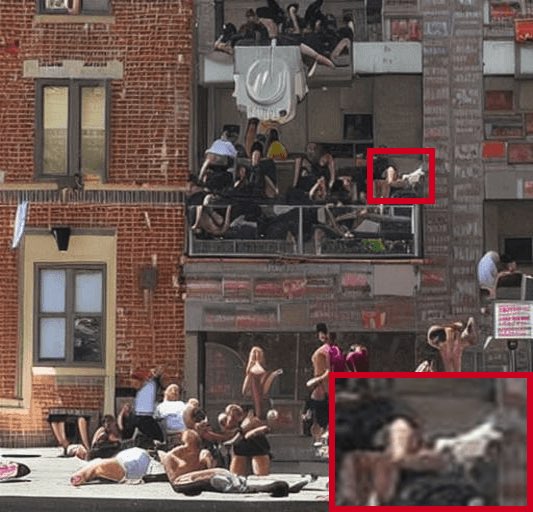}&
\includegraphics[width=0.116\linewidth,height=0.085\linewidth]{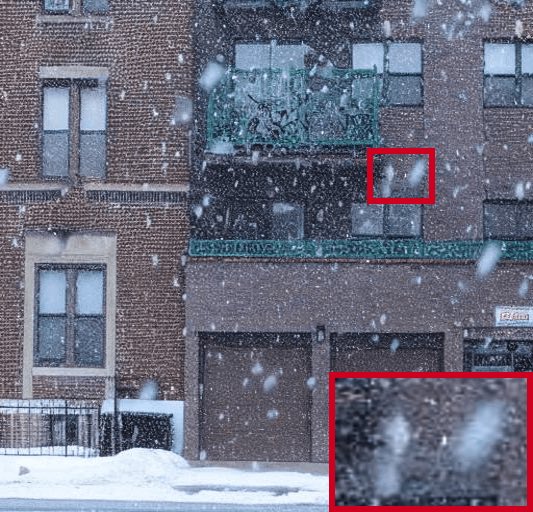}&
\includegraphics[width=0.116\linewidth,height=0.085\linewidth]{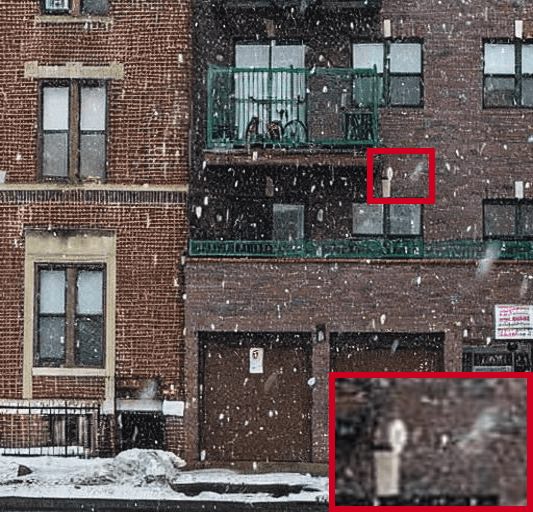}&
\includegraphics[width=0.116\linewidth,height=0.085\linewidth]{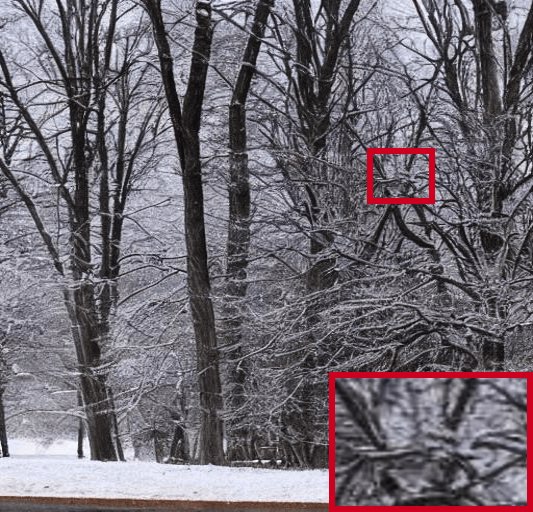}
\vspace{0mm}\\
\multirow{1}{*}[1cm]{\rotatebox{90}{\fontsize{8pt}{\baselineskip}\selectfont Dehazing}}
&\includegraphics[width=0.116\linewidth,height=0.085\linewidth]{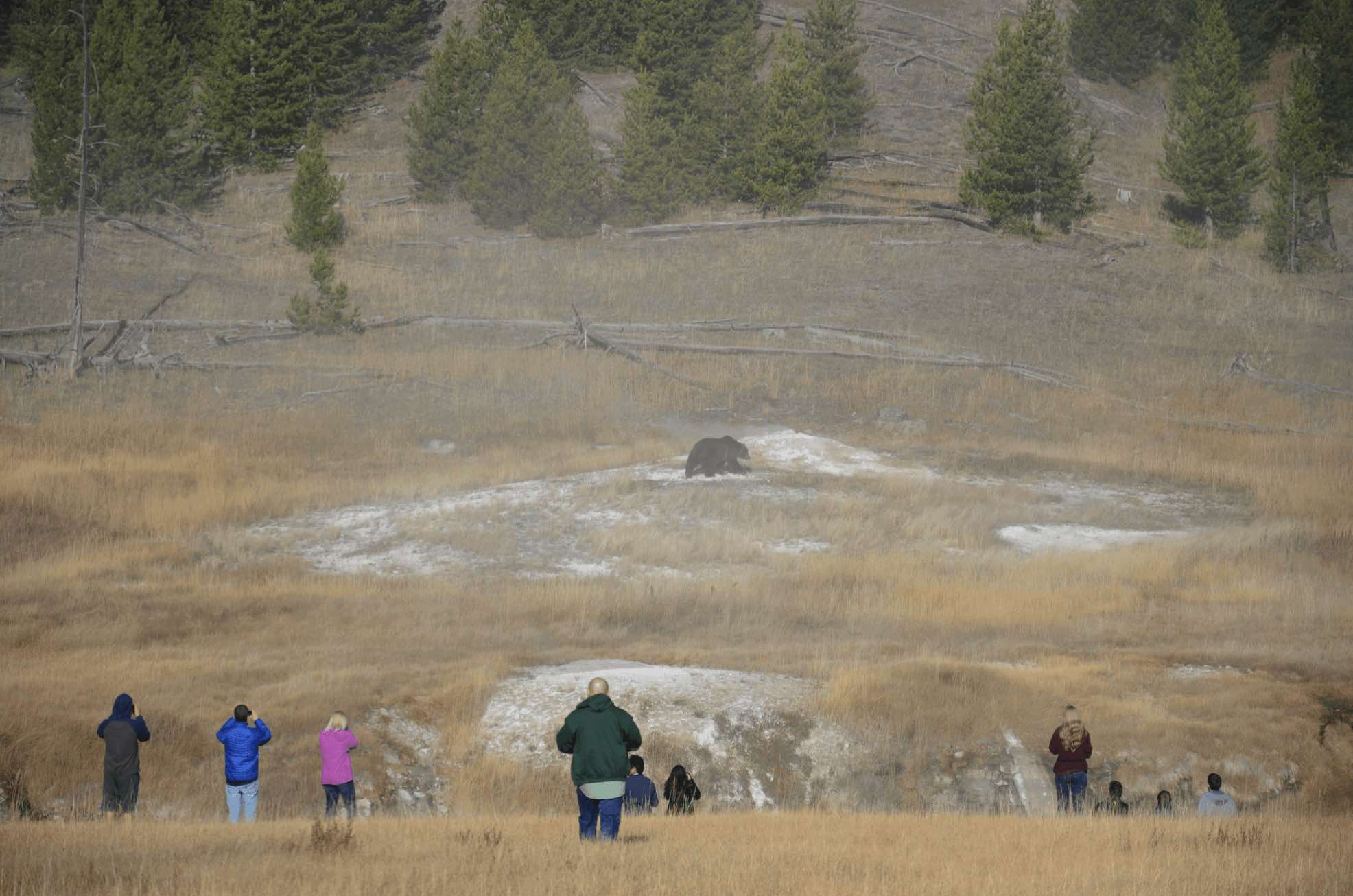}&
\includegraphics[width=0.116\linewidth,height=0.085\linewidth]{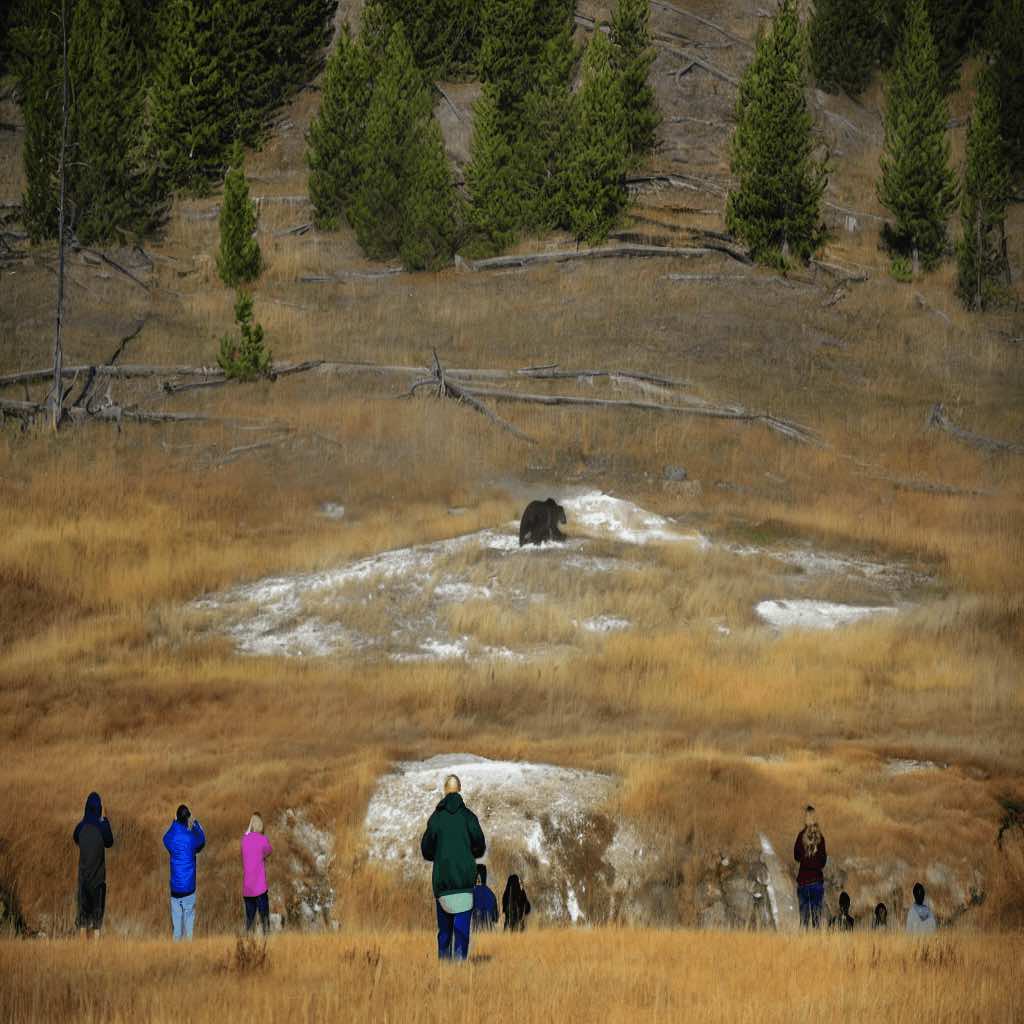}&
\includegraphics[width=0.116\linewidth,height=0.085\linewidth]{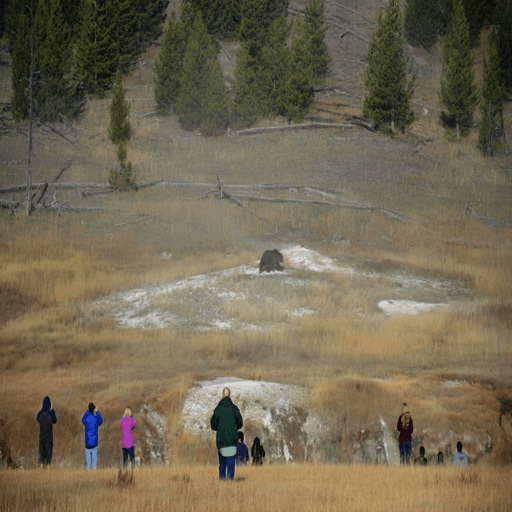}&
\includegraphics[width=0.116\linewidth,height=0.085\linewidth]{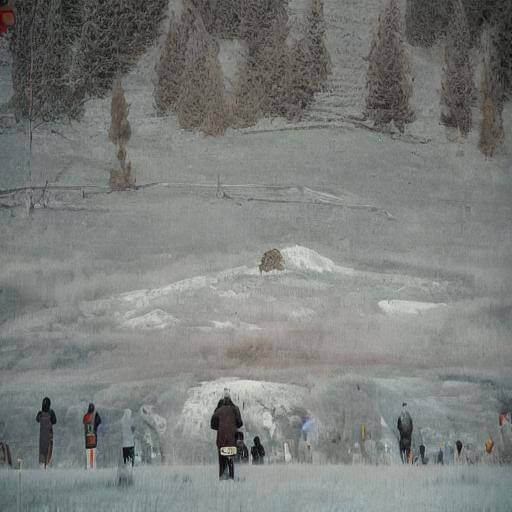}&
\includegraphics[width=0.116\linewidth,height=0.085\linewidth]{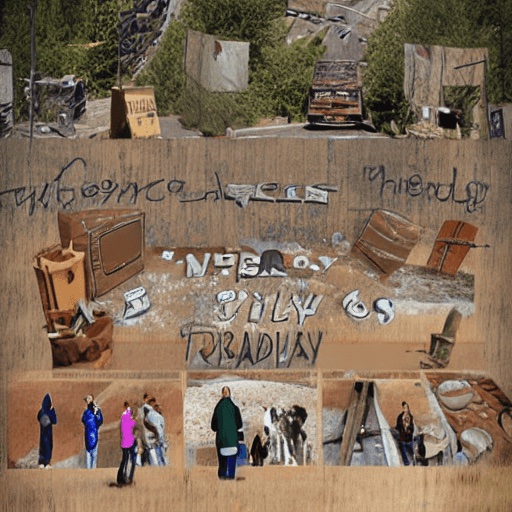}&
\includegraphics[width=0.116\linewidth,height=0.085\linewidth]{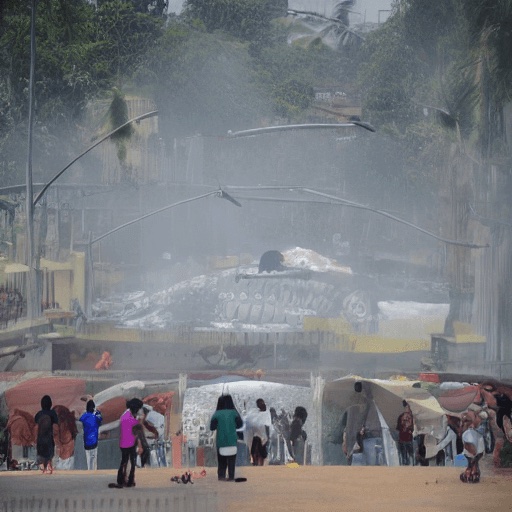}&
\includegraphics[width=0.116\linewidth,height=0.085\linewidth]{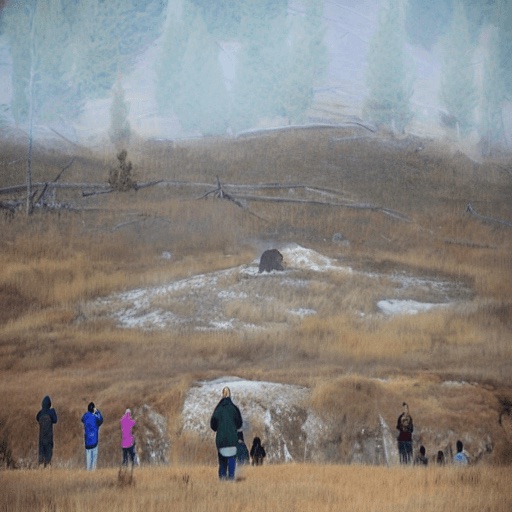}&
\includegraphics[width=0.116\linewidth,height=0.085\linewidth]{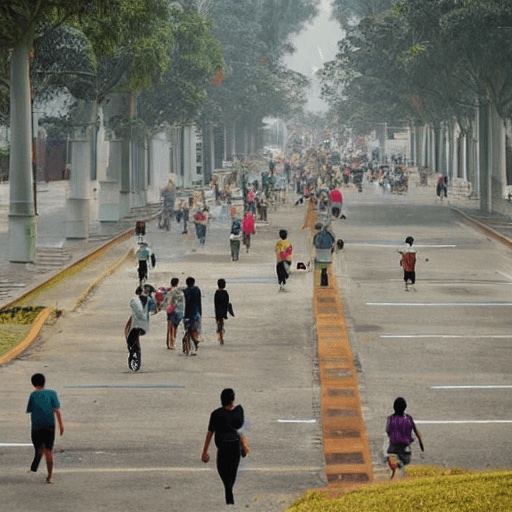}
\vspace{0mm}\\
\multirow{1}{*}[1.1cm]{\rotatebox{90}{\fontsize{8pt}{\baselineskip}\selectfont Deblurring}}
&\includegraphics[width=0.116\linewidth,height=0.085\linewidth]{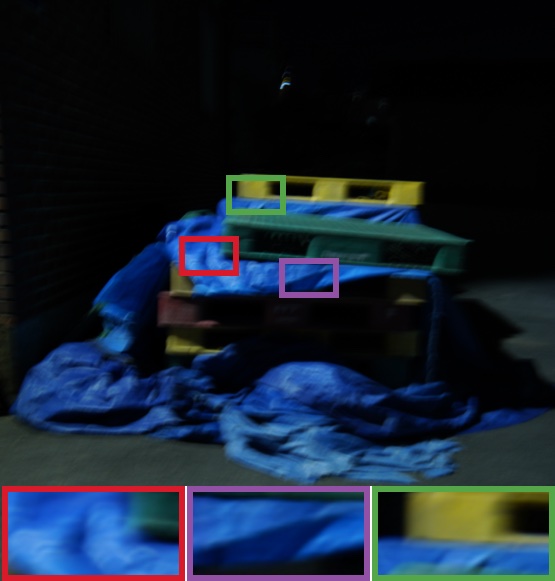}&
\includegraphics[width=0.116\linewidth,height=0.085\linewidth]{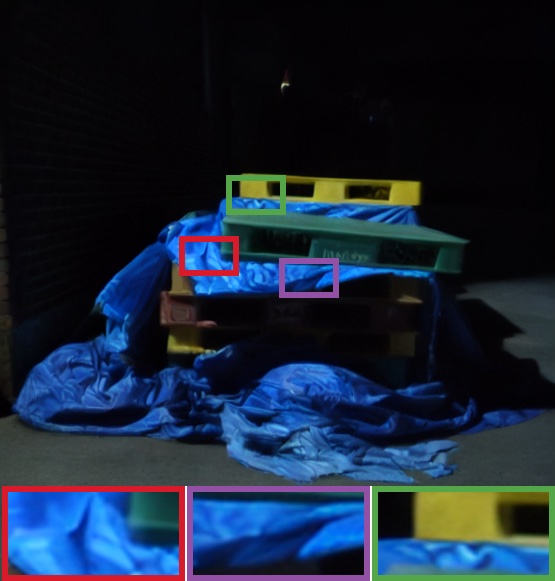}&
\includegraphics[width=0.116\linewidth,height=0.085\linewidth]{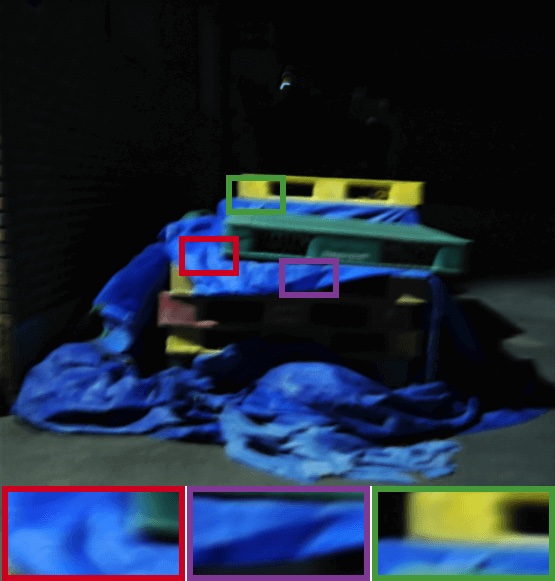}&
\includegraphics[width=0.116\linewidth,height=0.085\linewidth]{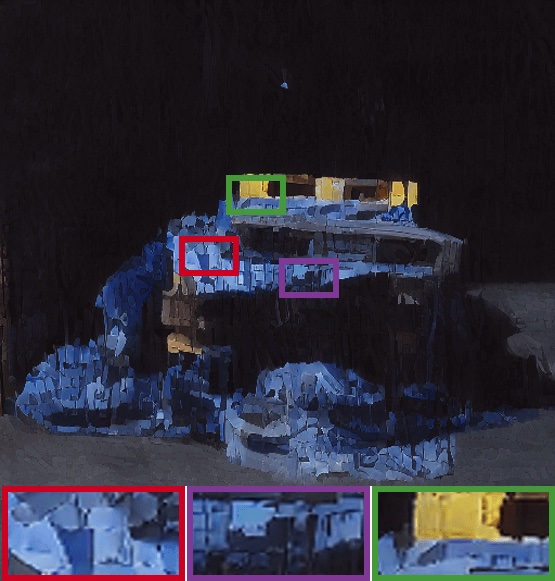}&
\includegraphics[width=0.116\linewidth,height=0.085\linewidth]{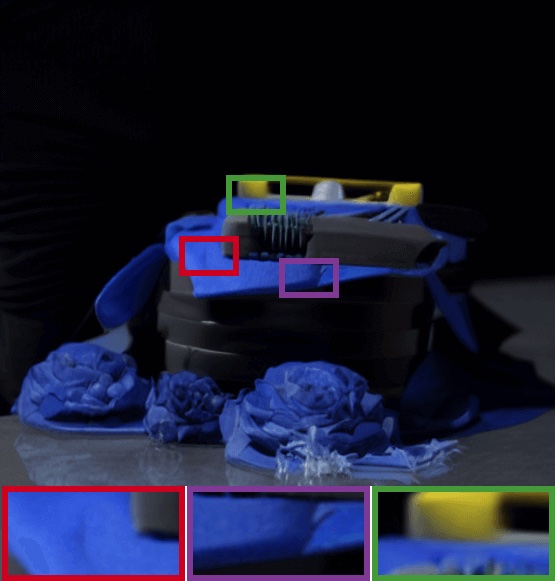}&
\includegraphics[width=0.116\linewidth,height=0.085\linewidth]{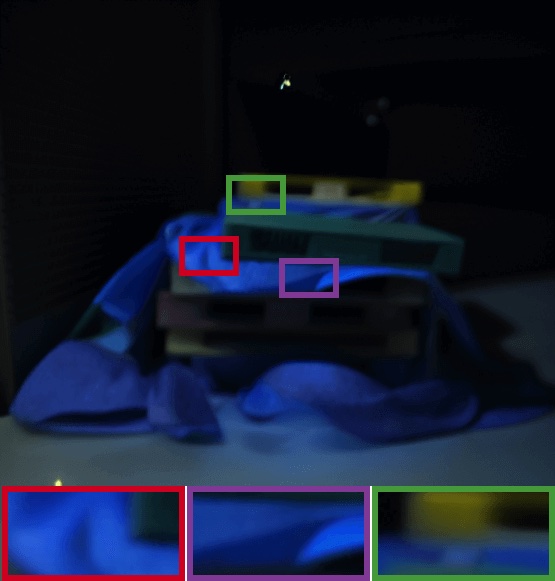}&
\includegraphics[width=0.116\linewidth,height=0.085\linewidth]{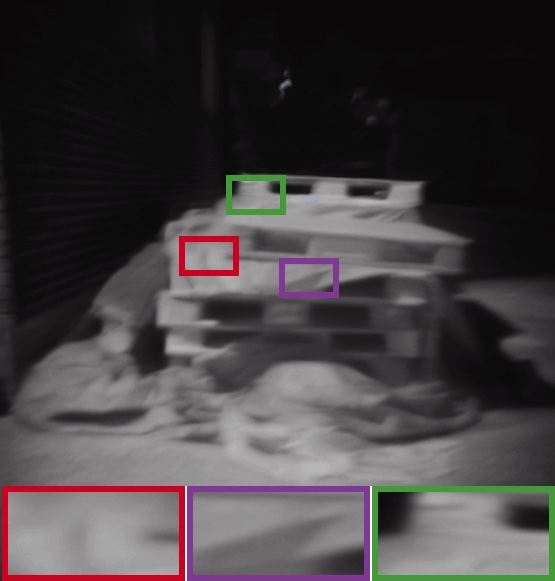}&
\includegraphics[width=0.116\linewidth,height=0.085\linewidth]{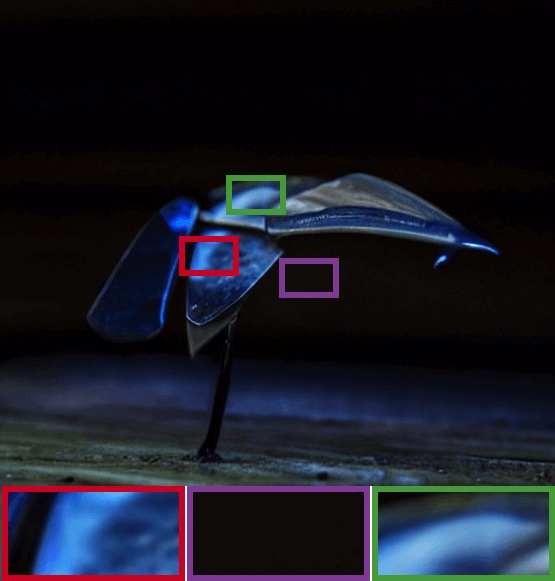}
\vspace{0mm}\\
\multirow{1}{*}[1cm]{\rotatebox{90}{\fontsize{8pt}{\baselineskip}\selectfont Deraining}}
&\includegraphics[width=0.116\linewidth,height=0.085\linewidth]{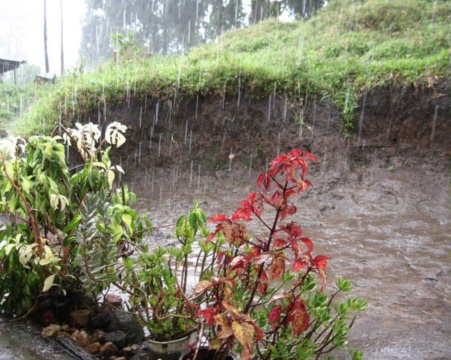}&
\includegraphics[width=0.116\linewidth,height=0.085\linewidth]{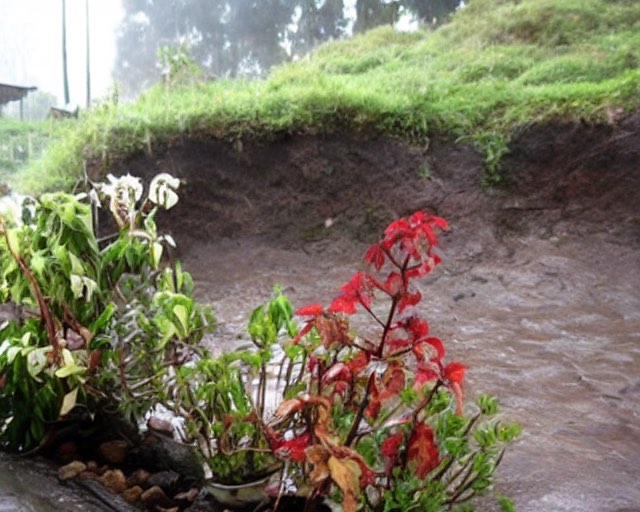}&
\includegraphics[width=0.116\linewidth,height=0.085\linewidth]{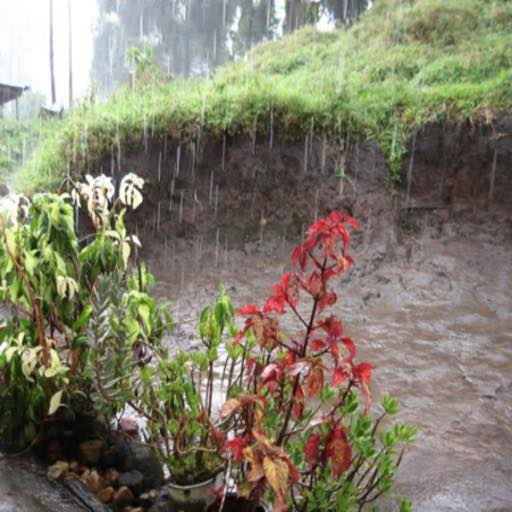}&
\includegraphics[width=0.116\linewidth,height=0.085\linewidth]{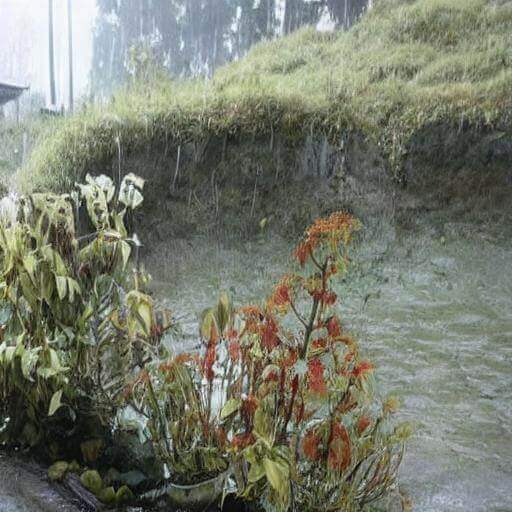}&
\includegraphics[width=0.116\linewidth,height=0.085\linewidth]{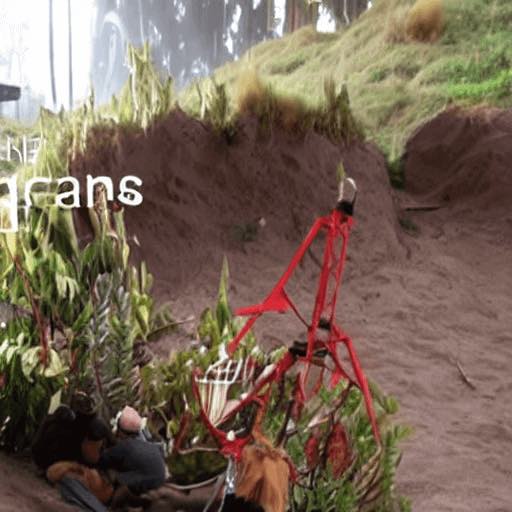}&
\includegraphics[width=0.116\linewidth,height=0.085\linewidth]{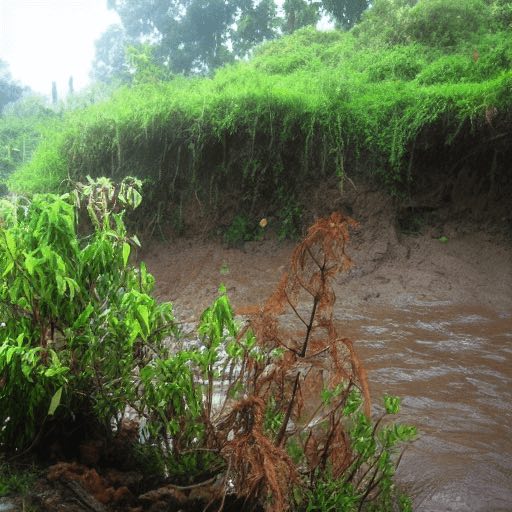}&
\includegraphics[width=0.116\linewidth,height=0.085\linewidth]{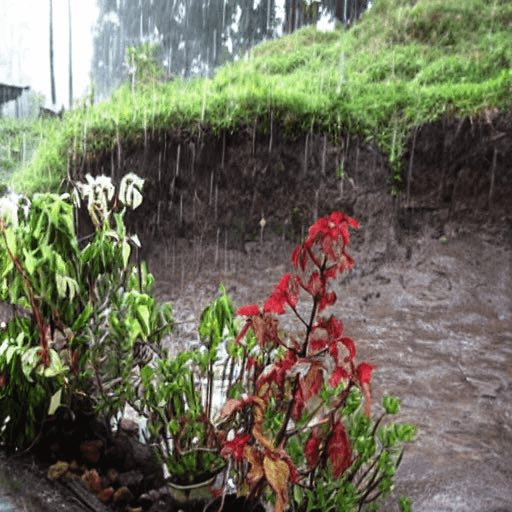}&
\includegraphics[width=0.116\linewidth,height=0.085\linewidth]{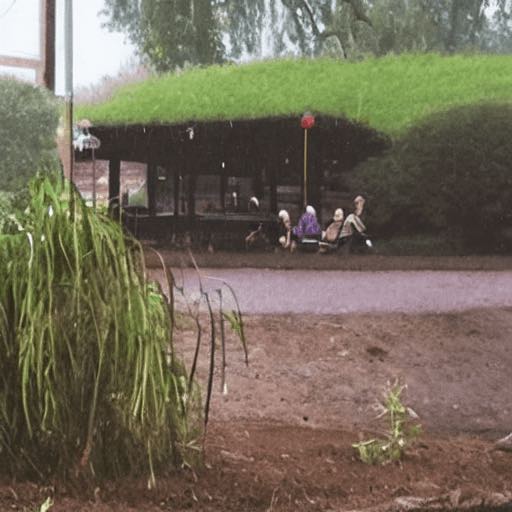}
\vspace{0mm}\\
\multirow{1}{*}[1.67cm]{\rotatebox{90}{\fontsize{8pt}{\baselineskip}\selectfont Face Restoration}}
&\includegraphics[width=0.116\linewidth,height=0.116\linewidth]{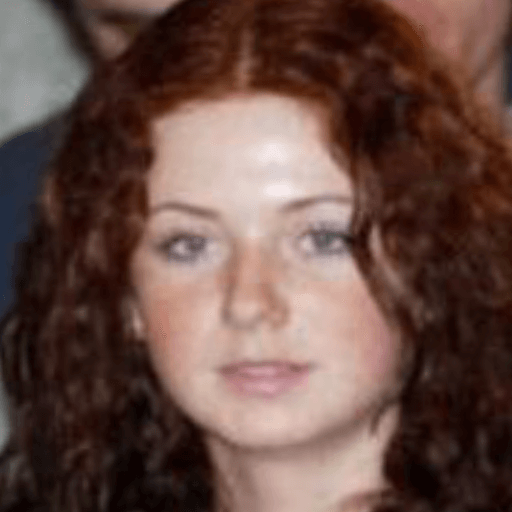}&
\includegraphics[width=0.116\linewidth,height=0.116\linewidth]{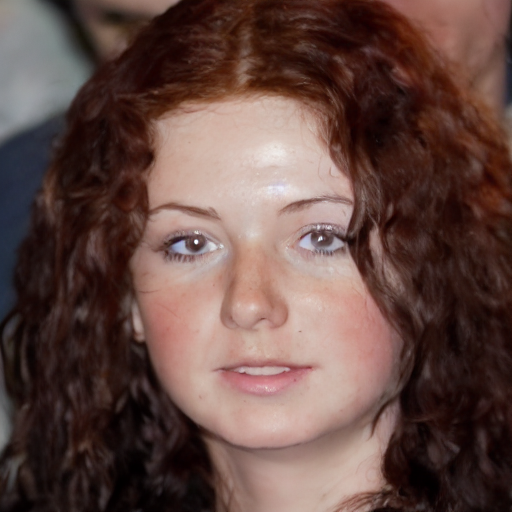}&
\includegraphics[width=0.116\linewidth,height=0.116\linewidth]{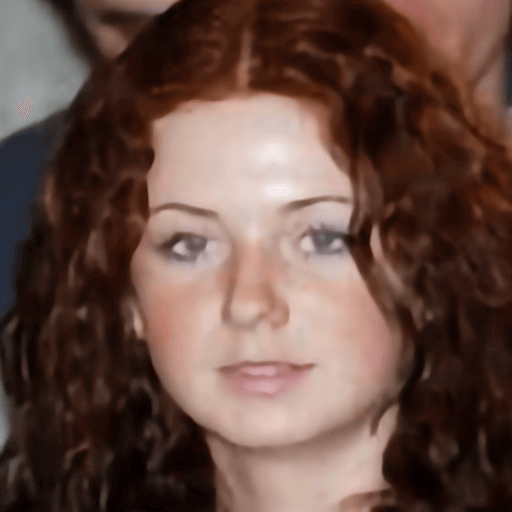}&
\includegraphics[width=0.116\linewidth,height=0.116\linewidth]{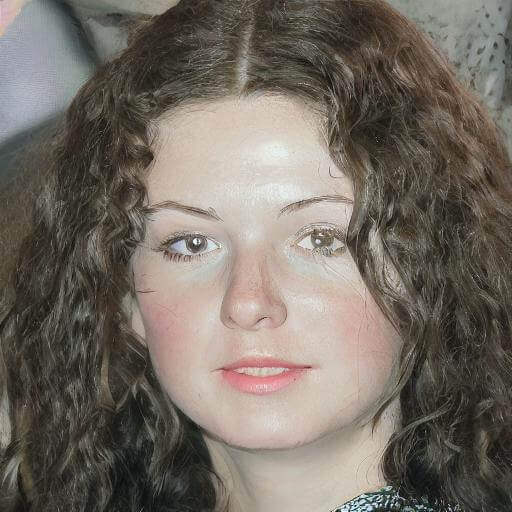}&
\includegraphics[width=0.116\linewidth,height=0.116\linewidth]{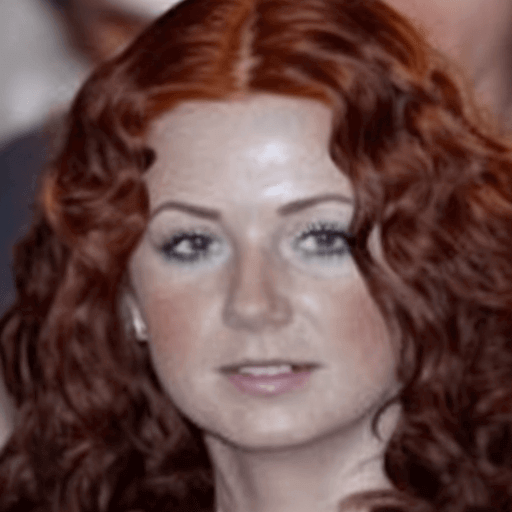}&
\includegraphics[width=0.116\linewidth,height=0.116\linewidth]{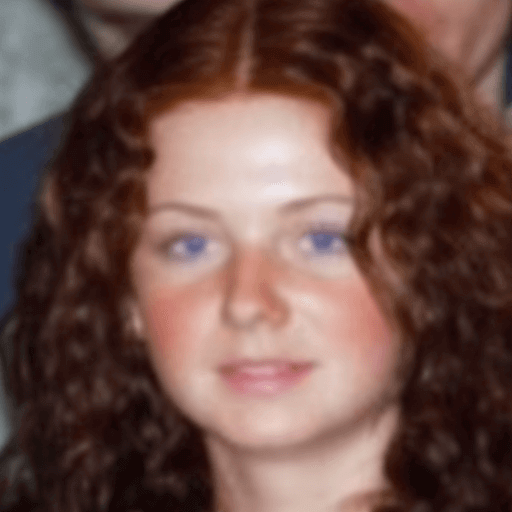}&
\includegraphics[width=0.116\linewidth,height=0.116\linewidth]{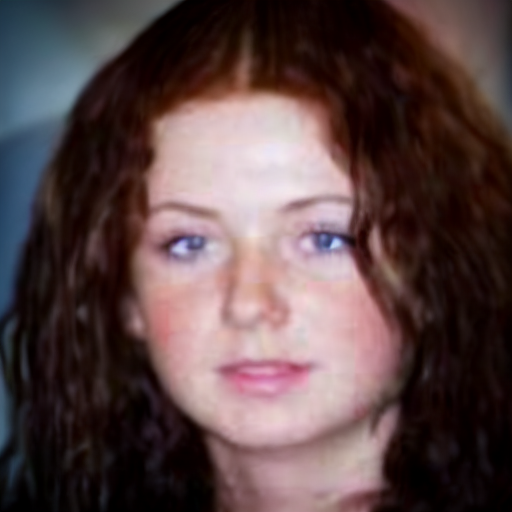}&
\includegraphics[width=0.116\linewidth,height=0.116\linewidth]{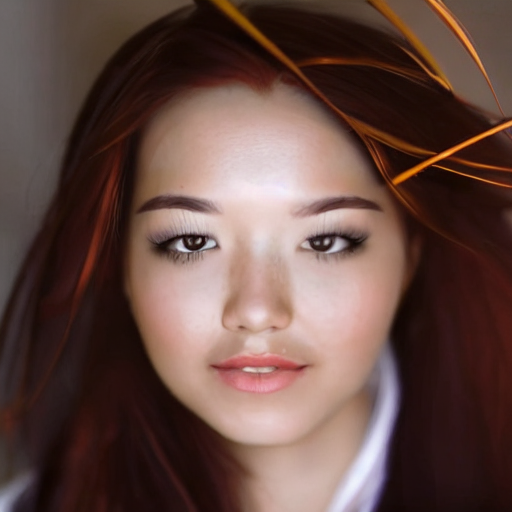}
\vspace{0mm}\\
\multirow{1}{*}[1.24cm]{\rotatebox{90}{\fontsize{8pt}{\baselineskip}\selectfont Low-light En.}}
&\includegraphics[width=0.116\linewidth,height=0.085\linewidth]{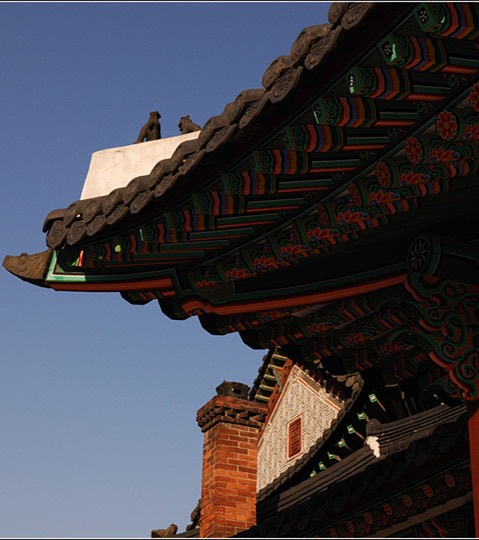}&
\includegraphics[width=0.116\linewidth,height=0.085\linewidth]{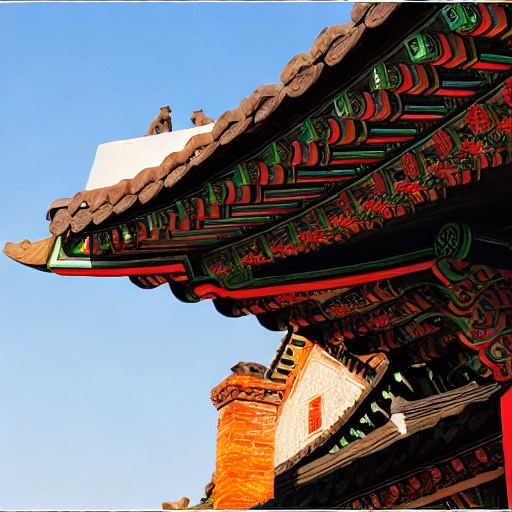}&
\includegraphics[width=0.116\linewidth,height=0.085\linewidth]{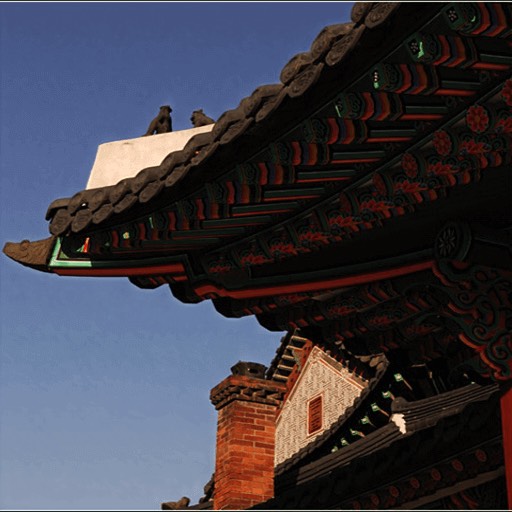}&
\includegraphics[width=0.116\linewidth,height=0.085\linewidth]{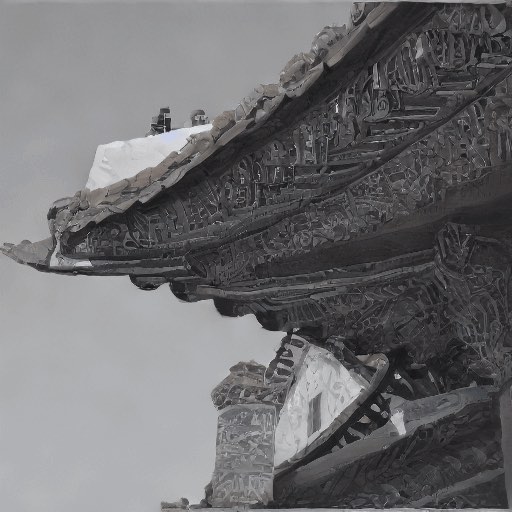}&
\includegraphics[width=0.116\linewidth,height=0.085\linewidth]{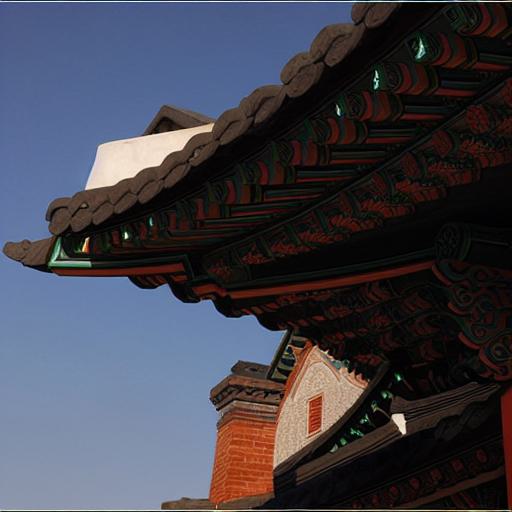}&
\includegraphics[width=0.116\linewidth,height=0.085\linewidth]{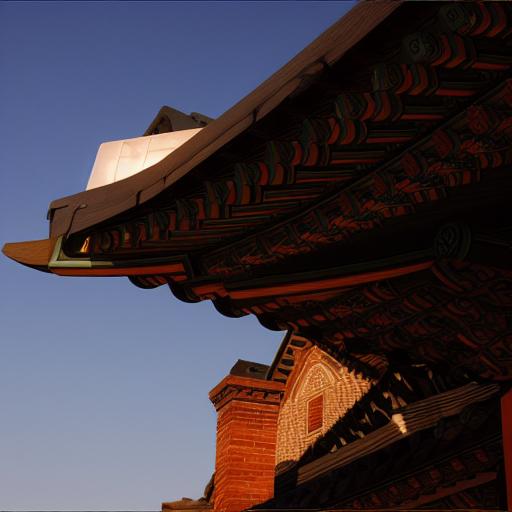}&
\includegraphics[width=0.116\linewidth,height=0.085\linewidth]{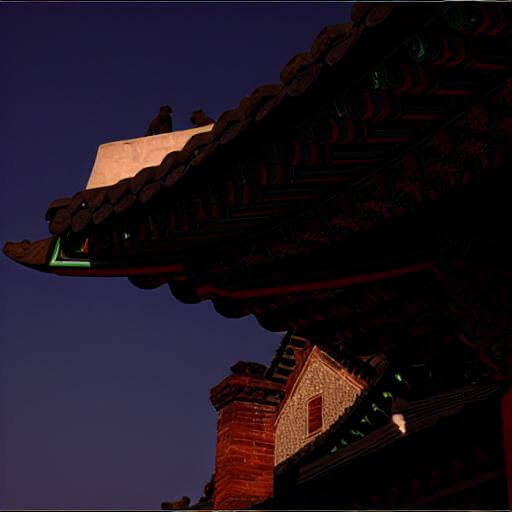}&
\includegraphics[width=0.116\linewidth,height=0.085\linewidth]{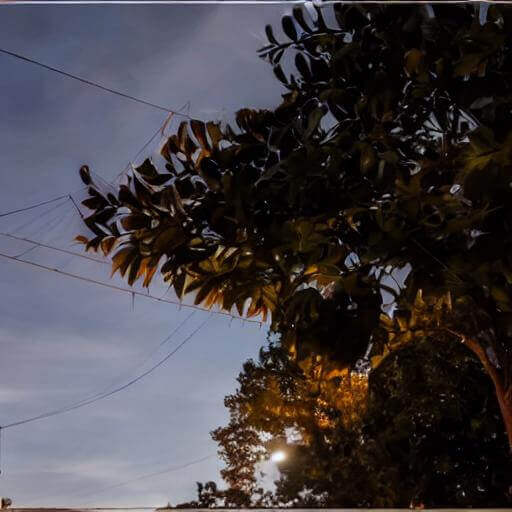}
\vspace{0mm}\\
\multirow{1}{*}[1.1cm]{\rotatebox{90}{\fontsize{8pt}{\baselineskip}\selectfont Demoireing}}
&\includegraphics[width=0.116\linewidth,height=0.085\linewidth]{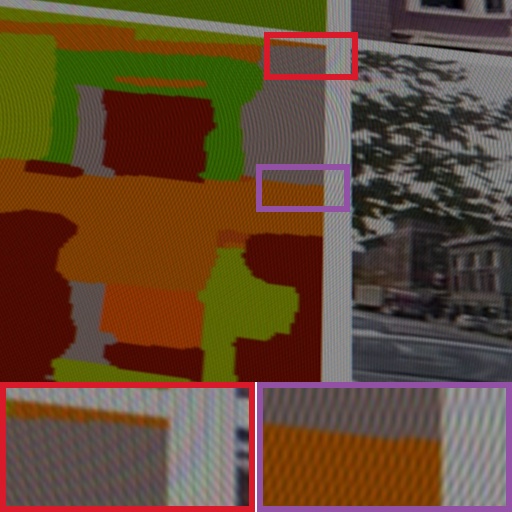}&
\includegraphics[width=0.116\linewidth,height=0.085\linewidth]{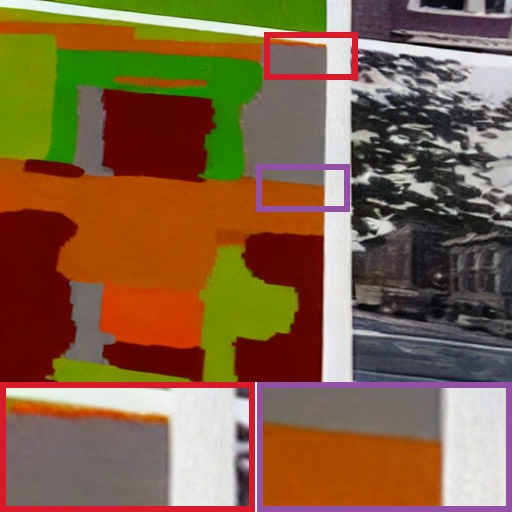}&
\includegraphics[width=0.116\linewidth,height=0.085\linewidth]{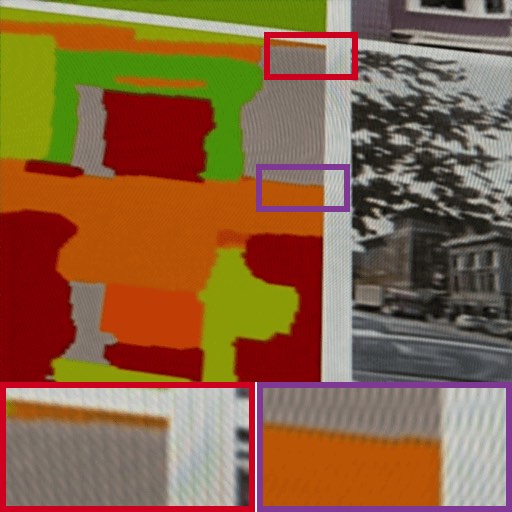}&
\includegraphics[width=0.116\linewidth,height=0.085\linewidth]{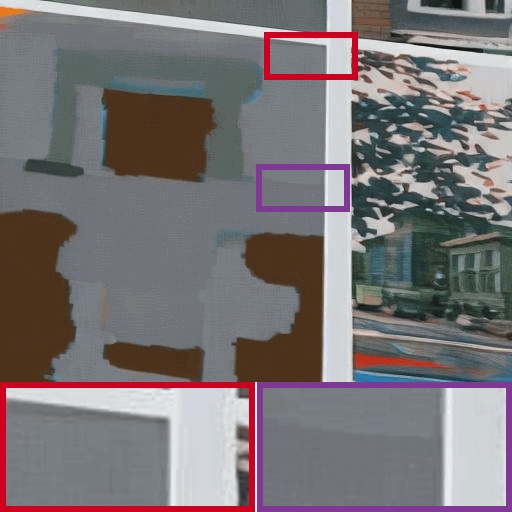}&
\includegraphics[width=0.116\linewidth,height=0.085\linewidth]{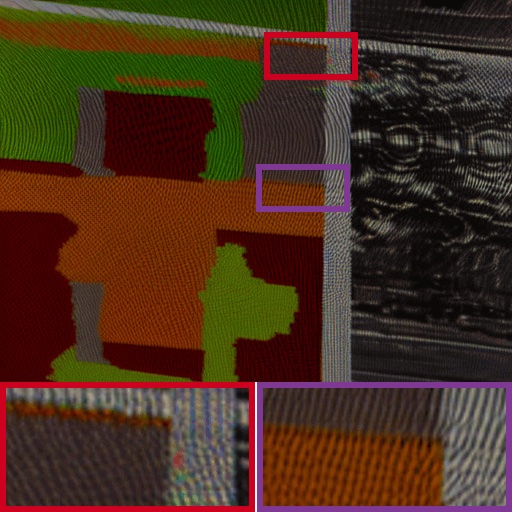}&
\includegraphics[width=0.116\linewidth,height=0.085\linewidth]{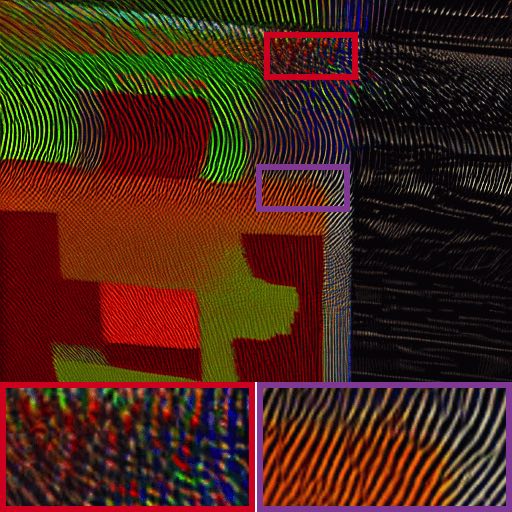}&
\includegraphics[width=0.116\linewidth,height=0.085\linewidth]{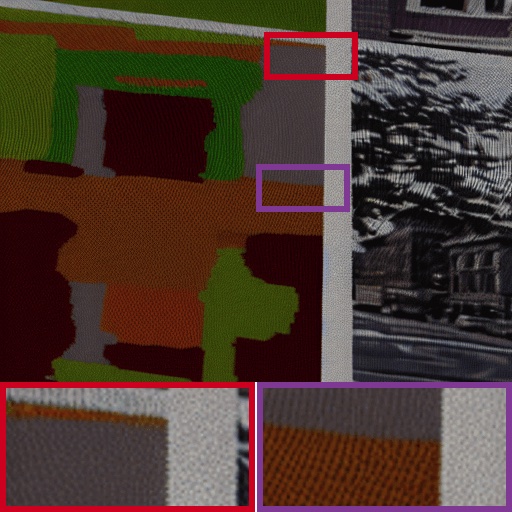}&
\includegraphics[width=0.116\linewidth,height=0.085\linewidth]{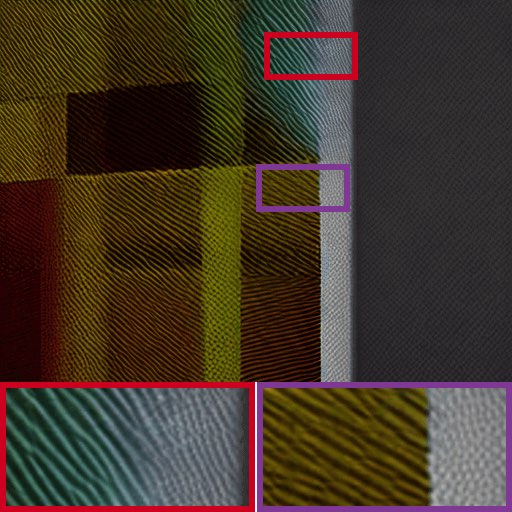}
\vspace{0mm}\\
\multirow{1}{*}[1.2cm]{\rotatebox{90}{\fontsize{8pt}{\baselineskip}\selectfont Highlight Re.}}
&\includegraphics[width=0.116\linewidth,height=0.085\linewidth]{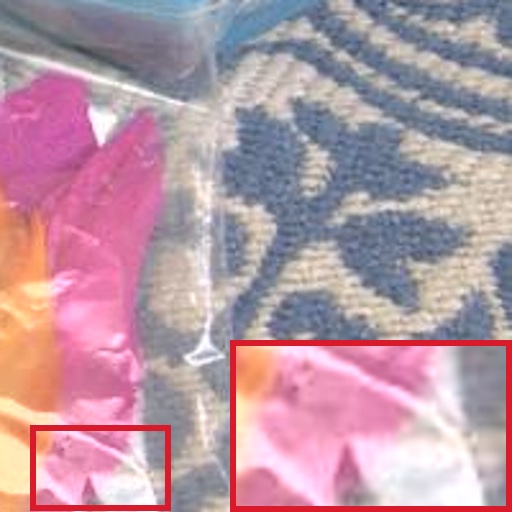}&
\includegraphics[width=0.116\linewidth,height=0.085\linewidth]{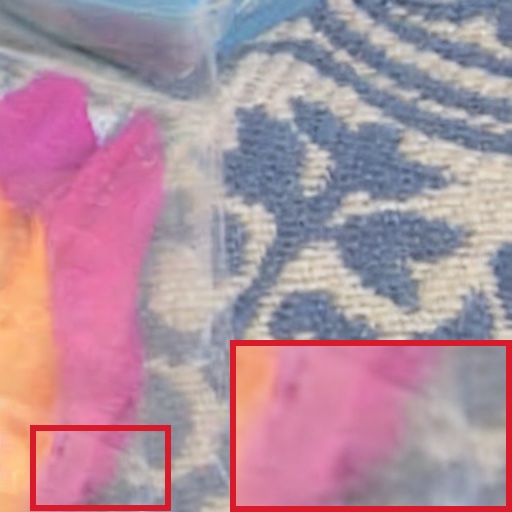}&
\includegraphics[width=0.116\linewidth,height=0.085\linewidth]{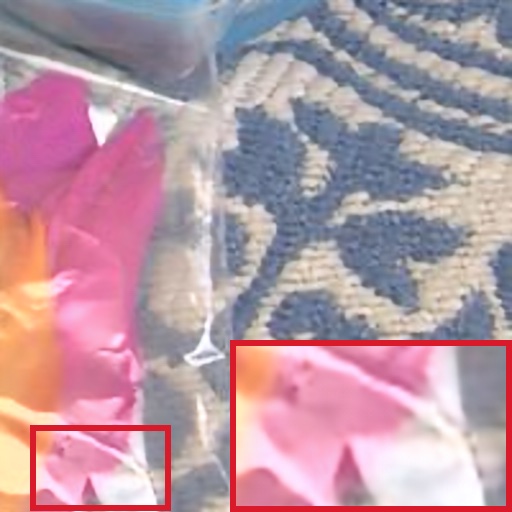}&
\includegraphics[width=0.116\linewidth,height=0.085\linewidth]{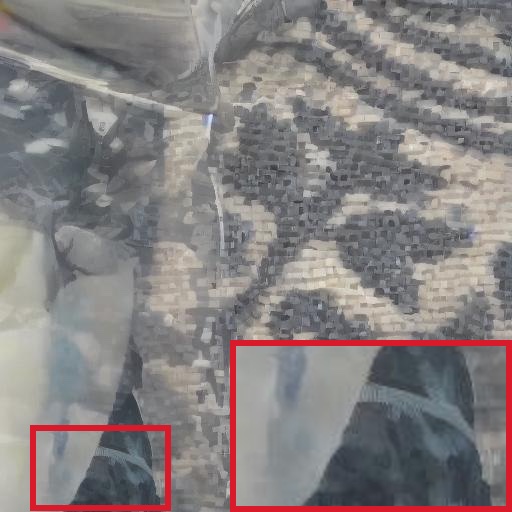}&
\includegraphics[width=0.116\linewidth,height=0.085\linewidth]{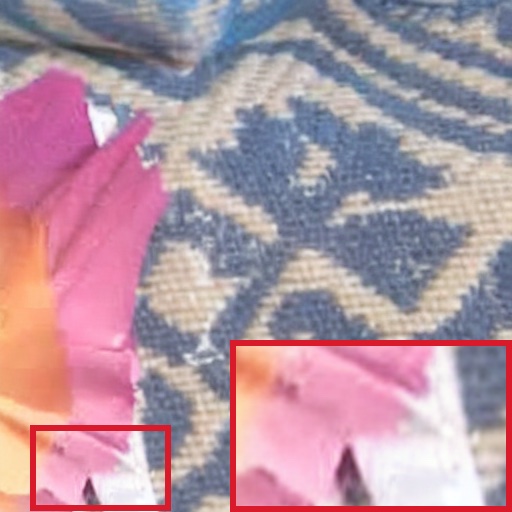}&
\includegraphics[width=0.116\linewidth,height=0.085\linewidth]{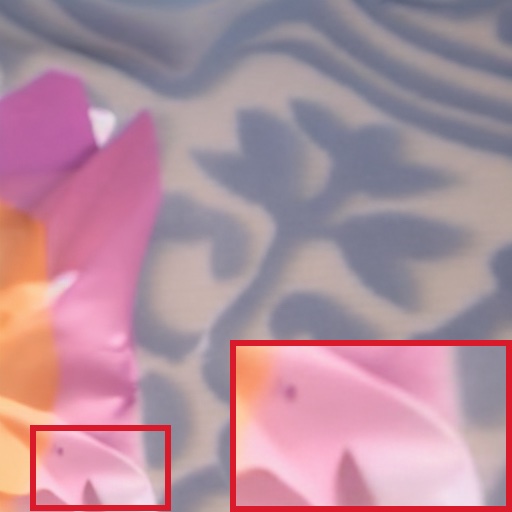}&
\includegraphics[width=0.116\linewidth,height=0.085\linewidth]{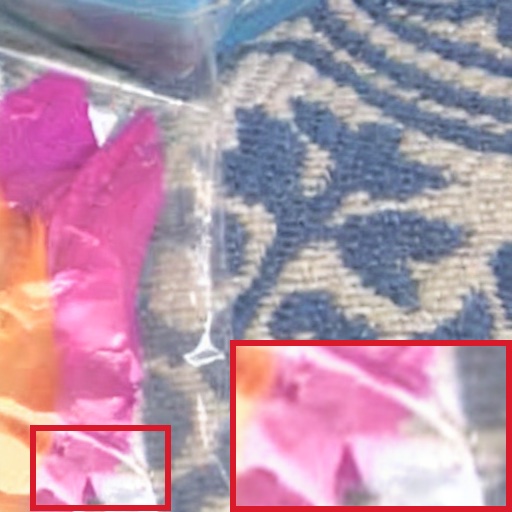}&
\includegraphics[width=0.116\linewidth,height=0.085\linewidth]{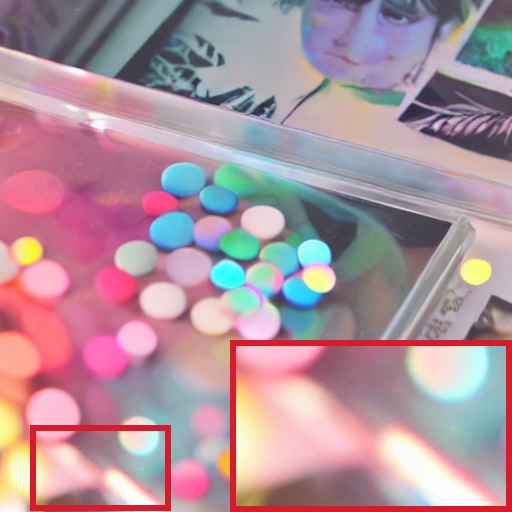}
\vspace{-1mm}\\
&\fontsize{7pt}{\baselineskip}\selectfont  (1) Input &
\fontsize{7pt}{\baselineskip}\selectfont  (2) Ours &
\fontsize{7pt}{\baselineskip}\selectfont (3) PromptIR \cite{potlapalli2023promptir}  &
\fontsize{7pt}{\baselineskip}\selectfont (4) ControlNet \cite{zhang2023adding} &
\fontsize{7pt}{\baselineskip}\selectfont (5) Null-Text \cite{mokady2023null} &
\fontsize{7pt}{\baselineskip}\selectfont (6) PNP \cite{tumanyan2023plug} &
\fontsize{7pt}{\baselineskip}\selectfont (7) InstructP2P \cite{brooks2023instructpix2pix} &
\fontsize{7pt}{\baselineskip}\selectfont (8) SD \cite{rombach2022high} 
\\
\end{tabular}
\end{center}
\vspace{-7mm}
\caption{\textbf{Qualitative Comparison.} Our \name notably surpasses regression-based method \yu{(3)} and diffusion-based methods \yu{(4)-(8)}   
in performance. \ryn{Magnified regions} of several tasks are provided for clarity. Refer to \textbf{Supplemental} for further comparisons.}
\label{fig:results}
\end{figure*}

\begin{table*}[t]
\centering
\tablestyle{5.5pt}{1.15}
\scalebox{0.77}{
\begin{tabular}{l|cc|cc|cc|cc|cc|cc|cc|cc}
\shline
 & \multicolumn{2}{c|}{Desnowing} &  
 \multicolumn{2}{c|}{Dehazing}  &  
 \multicolumn{2}{c|}{Deblurring} &
 \multicolumn{2}{c|}{Deraining} &  
 \multicolumn{2}{c|}{Low-light Enhanc.} &
 \multicolumn{2}{c|}{Face Restoration} &
 \multicolumn{2}{c|}{Demoireing} &
 \multicolumn{2}{c}{Highlight Removal}\\
& \multicolumn{2}{c|}{Realistic \cite{liu2018desnownet}}   
& \multicolumn{2}{c|}{Reside \cite{li2018benchmarking}} 
& \multicolumn{2}{c|}{RealBlur-J \cite{rim2020real}} 
& \multicolumn{2}{c|}{real test \cite{wang2019spatial}}
& \multicolumn{2}{c|}{merged low.} 
& \multicolumn{2}{c|}{LFW \cite{wang2021towards}} 
& \multicolumn{2}{c|}{LCDMoire \cite{yuan2019aim}} 
& \multicolumn{2}{c}{SHIQ \cite{fu2021multi}} \\
& FID $\downarrow$  & KID $\downarrow$ & FID $\downarrow$ & KID $\downarrow$ & FID $\downarrow$ & KID $\downarrow$ & FID $\downarrow$  & KID $\downarrow$ & FID $\downarrow$ & KID $\downarrow$ & FID $\downarrow$ & KID $\downarrow$ & FID $\downarrow$ & KID $\downarrow$ & FID $\downarrow$ & KID $\downarrow$ \\
\shline
\multicolumn{17}{c}{\small{\demphs{Regression-based \textit{specialized} models}}} \\
\hline
{\demphs{\textit{All}}}  &\demphs{33.92} &\demphs{5.39} &\demphs{36.40} &\demphs{15.66} &\demphs{55.64} &\demphs{15.70} &\demphs{52.78} &\demphs{16.28} &\demphs{48.47} &\demphs{10.96}  &\demphs{19.28} &\demphs{6.72}  &\demphs{29.59} &\demphs{1.45}  &\demphs{33.74} &\demphs{18.79}  \\
\shline
\multicolumn{17}{c}{\small{Regression-based \textit{multi-task} models}} \\
\hline
AirNet* \cite{li2022all} &35.02  &5.52  &39.53  &17.86  &59.38  &20.95  &52.04  &16.20  &\fang{59.92}  &\fang{19.74}  &31.03  &13.35  &33.05  &\fang{4.27}  &\fang{10.13}  &\fang{5.89}     \\
WGWS-Net* \cite{zhu2023learning}  &34.84  &5.71  &36.25  &15.79  &56.80  &16.83 &53.64 &16.55 &53.67  &12.99  &29.89  &12.08  &29.86 &2.28  &\hlcolor{8.28}  &\hlcolor{3.05}    \\
PromptIR* \cite{potlapalli2023promptir} &34.66 &5.35  &40.88 &17.80  &55.37  &16.42 &53.78 &16.88  &53.42  &13.16 &30.52 &12.80  &\hlcolor{29.01} &\hlcolor{1.56}  &\hlcolor[second]{9.01}  &\hlcolor[second]{5.07}     \\ 
\hline
\multicolumn{17}{c}{\small{{Diffusion-based models}}} \\
\hline
SD \cite{rombach2022high} & 35.24 & 7.88 & \fang{48.89} &\fang{24.47} & 59.21 & 18.96 & 51.78 & \fang{17.69} & \fang{53.09} &\fang{15.38} &30.90 &\fang{9.63} & 58.20 & 17.34 &\fang{36.54} &\fang{12.06}    \\ 
PNP \cite{xiao2023plug} &35.01 &\fang{6.52}&\fang{42.82}&\fang{16.98}&63.16&23.58  &52.89 &21.02 &\fang{54.19} &\fang{14.43}  &\fang{34.08}&\fang{13.45} &\fang{36.37} &\fang{6.18} &\fang{33.09} &14.94\\
P2P \cite{parmar2023zero}&34.48 &6.03 &42.17&17.33&63.43&25.15  &\hlcolor{44.49}&\hlcolor[second]{13.94} & \fang{52.06} &\fang{13.26}  & 54.67 & 24.66&36.37&9.35&26.96&13.11   \\
InstructP2P \cite{brooks2023instructpix2pix}&42.01 &\fang{8.54}&\hlcolor{33.48}&\hlcolor{12.76}&57.38 &\fang{19.37}  &54.12 &\fang{17.87} &\fang{55.65}  &\fang{15.25}  &\fang{24.66} &\fang{9.73} &\fang{34.29} &4.73 &\fang{16.80}  &\fang{6.81}    \\ 
Null-Text \cite{mokady2023null}&60.49&16.38 &39.94  &14.88 &60.38&20.37 &\fang{51.49}&15.43&\fang{52.86} &\hlcolor[second]{12.79} &\fang{33.06} &\fang{12.82} &33.72 &\fang{4.91} &\fang{14.65} &\fang{6.52} \\
ControlNet* \cite{zhang2023adding}&\hlcolor[second]{34.36} &\hlcolor[second]{5.70} &37.02&15.45&\hlcolor[second]{52.30}&\hlcolor[second]{17.19} &52.55&15.22&\hlcolor[second]{51.56} &\fang{15.51}  &\hlcolor[second]{21.59}&\hlcolor[second]{7.84}&41.97&8.80&15.75&8.17  \\ 
Diff-Plugin (ours)  & \hlcolor{34.30} & \hlcolor{5.20} & \hlcolor[second]{34.68} & \hlcolor[second]{14.38} & \hlcolor{51.81} & \hlcolor{14.63}   & \hlcolor[second]{50.55} & \hlcolor{13.84} & \hlcolor{48.98}  & \hlcolor{11.73}   & \hlcolor{20.07} & \hlcolor{6.91} & \hlcolor[second]{29.77} & \hlcolor[second]{1.75} & 12.58 & 6.37 \\
\shline
\end{tabular}
}
\vspace{-3mm}
\caption{Quantitative comparisons to SOTAs (both regression-based and diffusion-based methods) on eight low-level vision tasks that need high content-preservation. We summarise all the regression-based specialized models in one line, denoted as ``\textit{\demphs{All}}''. 
They are: 
DDMSNet \cite{zhang2021deep} (desnowing), 
PMNet \cite{ye2022perceiving} (dehazing), 
Restormer \cite{zamir2022restormer} (deblurring and deraining), 
NeRCO \cite{yang2023implicit} (low-light enhancement), 
VQFR \cite{gu2022vqfr} (face restoration),
UHDM \cite{yu2022towards} (demoireing), 
SHIQ \cite{fu2021multi} (highlight removal).
KID values are scaled by a factor of 100 for readability. 
* means that this method is re-trained on eight tasks by us. The \hlcolor{best} and \hlcolor[second]{second-best} results are highlighted.}
\label{tab:comp_sota}
\end{table*}

\begin{table*}[t!]
\centering
\footnotesize
\setlength{\tabcolsep}{3pt}
\renewcommand\arraystretch{1.}
\begin{tabular}{l|cccccccccccc}
    \shline
     Methods &AirNet \cite{li2022all} &WGWS-Net \cite{zhu2023learning}  &PromptIR \cite{potlapalli2023promptir}  
     &SD \cite{rombach2022high}
     &PNP \cite{tumanyan2023plug} &P2P \cite{parmar2023zero} &InstructP2P \cite{brooks2023instructpix2pix} &Null-Text \cite{mokady2023null} &ControlNet \cite{zhang2023adding}  & Ours  \\
    \hline
    AR $\downarrow$  &5.26 &\hlcolor[second]{2.75}   &3.04 &9.66  &6.32 &7.39  &7.14  &7.94  &4.33  &\hlcolor{1.17}  \\ 
    \shline
\end{tabular}
\vspace{-3mm}
\caption{Average Ranking (AR) of different methods in the User Study. 
The lower the value, the better the human subjective evaluation.
}
\label{tab:user_study}
\end{table*}

\section{Experiments}
\label{sec:experiments}

In this section, we first introduce our experimental setup, \ryn{including} datasets, implementation, and metrics. We then compare \name with current diffusion- and regression-based methods in \secref{subsec:comp_sota}, and conduct component analysis of \name via \ryn{ ablation studies} in \secref{subsec:ablation}.

\noindent \textbf{Datasets.} 
To train the Task-Plugins, we utilize specific datasets for each low-level task, 
desnowing: Snow100K \cite{liu2018desnownet}, 
dehazing: Reside \cite{li2018benchmarking}, 
deblurring: Gopro \cite{nah2017deep}, 
deraining: merged train \cite{zamir2021multi}, 
face restoration: FFHQ \cite{karras2019style}, 
low-light enhancement: LOL \cite{wei2018deep}, 
demoireing: LCDMoire \cite{yuan2019aim}, 
and highlight removal: SHIQ \cite{fu2021multi}. 
For testing, we evaluate on real-world benchmark datasets, 
desnowing: realistic test \cite{liu2018desnownet}, 
dehazing: RTTS \cite{li2018benchmarking}, 
deblurring: RealBlur-J \cite{rim2020real}, 
deraining: real test \cite{wang2019spatial}, 
face restoration: LFW \cite{huang2008labeled,wang2021towards}, 
low-light enhancement: merged low-light \cite{wei2018deep,wang2013naturalness,lee2013contrast,ma2015perceptual,guo2016lime,vonikakis2018evaluation}, 
demoireing: LCDMoire \cite{yuan2019aim}, 
and highlight removal: SHIQ \cite{fu2021multi}. 
To train the Plugin-Selector, we employ GPT \cite{openai2023gpt4} to generate text prompts for each task.

\noindent \textbf{Implementation.} 
During training and testing, we resize the image to 512$\times$512 for a fair comparison.  We employ the AdamW optimizer \cite{loshchilov2017decoupled} with its default parameters (\eg, betas, weight decay). 
The training of our Task-Plugins was conducted using a constant learning rate of $1e^{-5}$ and a batch size of 64 on four A100 GPUs, each with 80G of memory. 
To train the Plugin-Selector, we randomly sample 5,000 images from each task and augment text diversity by randomly combining text inputs from various tasks. We set the batch size to 8 and adopt the same learning rate for Task-Plugins. 
For negative texts, we set $N = 7$ by default. 
During inference, we set the specified similarity threshold $\theta =0$.

\noindent \textbf{Metrics.} 
We follow \cite{rombach2022high} to employ widely adopted non-reference perceptual metrics, FID \cite{heusel2017gans} and KID \cite{binkowski2018demystifying}, to evaluate our \textit{Diff-Plugin} on real data, as GT is not always available. 
As for the Plugin-Selector, we follow multi-label object classification \cite{chen2019multi} to report the mean average precision (mAP), the average per-class precision (CP), F1 (CF1), and the average overall precision (OP), recall (OR), and F1 (OF1). 
For each class (\ie, task type), the labels are predicted as positive if their confidence score is greater than $\theta$. \yu{We further propose a stringent zero-tolerance evaluation metric (ZTA) that rigorously assesses sentence-level classification results from a user-first perspective, making binary classification to ensure utmost accuracy:}
\begin{equation}
    \text{ZTA} = \frac{1}{Q} \sum_{i=1}^{Q} \left( \left( \min_{j \in Y_i} S_{ij} > \theta \right) \land \left( \max_{k \in H_i} S_{ik} \leq \theta \right) \right) \text{,}
    \label{eq:zta}
\end{equation}
where $Q$ is the total number of test samples, $S_i$ is the set of predicted similarity scores for sample $i$, $Y_i$ is the set of indices for positive classes (\ie, user interested tasks),  $H_i$ is the set of indices for negative classes (\ie, irrelevant tasks).

\subsection{Comparison with State-of-the-Art Methods}
\label{subsec:comp_sota}
We compare the proposed \textit{Diff-Plugin} with the state-of-the-art methods from different low-level vision tasks, including regression-based specialized models: 
DDMSNet \cite{zhang2021deep}, 
PMNet \cite{ye2022perceiving}, 
Restormer \cite{zamir2022restormer}, 
NeRCO \cite{yang2023implicit}, 
VQFR \cite{gu2022vqfr}, 
UHDM \cite{yu2022towards}, 
SHIQ \cite{fu2021multi}, 
multi-task models: AirNet \cite{li2022all}, 
WGWS-Net \cite{zhu2023learning} and
PromptIR \cite{potlapalli2023promptir}, 
and diffusion-based models: 
SD \cite{rombach2022high},
PNP \cite{tumanyan2023plug},
P2P \cite{parmar2023zero},
InstructP2P \cite{brooks2023instructpix2pix},
Null-Text \cite{mokady2023null}
and ControlNet \cite{zhang2023adding}.  We conduct the experiment on real-world datasets to compare the generalization \ryn{ability}. 

\noindent \textbf{Qualitative Results.} 
\figref{fig:results} \ryn{demonstrates the superior performances of our \textit{Diff-Plugin} on} eight low-level vision tasks with challenging natural images. 
First, using SD's \textit{img2img} \cite{rombach2022high} function does not ensure content accuracy. \ryn{It often leads} to major scene changes (column 8). 
InstructP2P \cite{brooks2023instructpix2pix}, \ryn{which lacks} task-specific priors, also falls short, producing poorer results in tasks like dehazing and low-light \ryn{enhancement} (column 7). 
The lack of task-specific priors also leads P2P \cite{parmar2023zero} and Null-Text \cite{mokady2023null} \ryn{into generating} inconsistent contents (columns 5 and 6), despite using \yu{initial noise from DDIM Inversion} \cite{song2020denoising}. ControlNet \cite{zhang2023adding} handles some tasks well (column 4) by providing condition information via a diffusion branch, but its strong color distortion reduces its effectiveness in \fang{these} tasks. 
The latest multi-task method, PromptIR  \cite{potlapalli2023promptir} (column 3), is limited by model scale and can only handle a few tasks. 
In contrast, our method uses a lightweight task-specific plugin for each task, offering flexibility and stable performance across all tasks (column 2). 

\noindent \textbf{Quantitative Results.} We also provide the quantitative comparison in \tableref{tab:comp_sota}. 
Compared with diffusion-based methods, our Diff-Plugin achieves SOTA results overall. 
\ryn{While PNP \cite{tumanyan2023plug} and InstructP2P \cite{brooks2023instructpix2pix} are capable of producing high-quality images with low FID \& KID, they often produce} significant content alterations (refer to \figref{fig:results}). 
Compared with regression-based multi-task methods, our approach delivers competitive performances in most tasks, though it is slightly ineffective 
in sparse degradation tasks like demoireing and highlight removal. 
While specialized models may outperform ours in their respective areas, their task-dependent designs limit their applicability to other tasks. 
\ryn{Note that the} primary goal of this paper is not to achieve top \ryn{performances in all} tasks, but to lay groundwork for future advancements. 
In addition, \textit{Diff-Plugin},  enables text-driven low-level task processing, a capability absent in regression-based models.

\noindent\textbf{User Study.} We conduct a user study with 46 participants to assess various methods through subjective evaluation. 
Each participant reviewed 5 image sets from the test set, each comprising an input image and 10 predicted images, for a total of 8 tasks. 
The images were ranked based on content consistency, degradation removal (\eg, rain, snow, highlight), and overall quality. 
Analyzing 1,840 rankings (46 participants $\times$ 40 sets), we compute the Average Ranking (AR) of each method. 
\tableref{tab:user_study} shows the results. It is obvious to see a preference for our approach among the users.

\begin{figure*}[t!]
\begin{center}
\scalebox{0.88}{
\begin{tabular}{c@{\hspace{2.8mm}}c@{\hspace{2.8mm}}c@{\hspace{2.8mm}}c@{\hspace{2.8mm}}c@{\hspace{2.8mm}}c@{\hspace{2.8mm}}c@{\hspace{2.8mm}}c@{\hspace{2.8mm}}c@{\hspace{2.8mm}}}
\includegraphics[width=0.14\linewidth,height=0.08\linewidth]{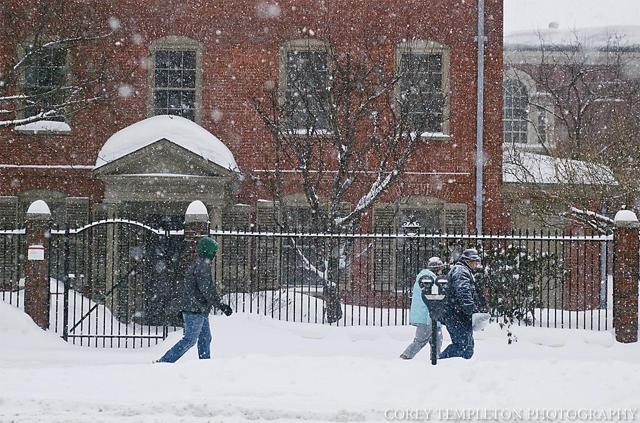}& 
\includegraphics[width=0.14\linewidth,height=0.08\linewidth]{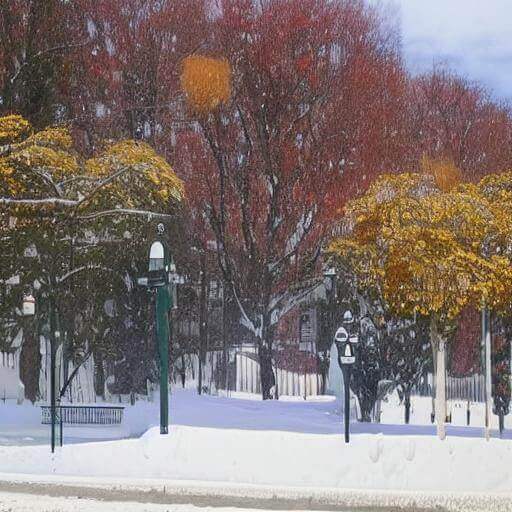}& 
\includegraphics[width=0.14\linewidth,height=0.08\linewidth]{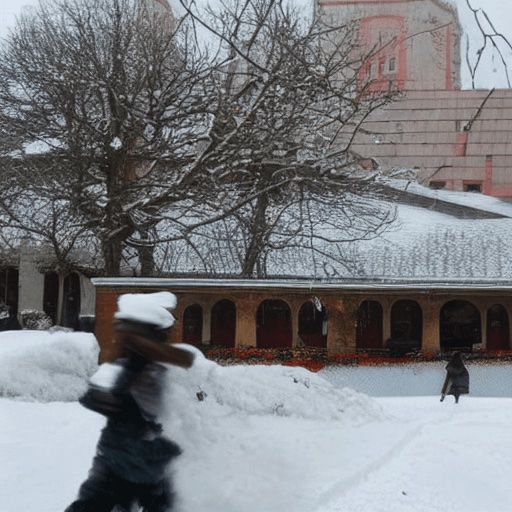}& 
\includegraphics[width=0.14\linewidth,height=0.08\linewidth]{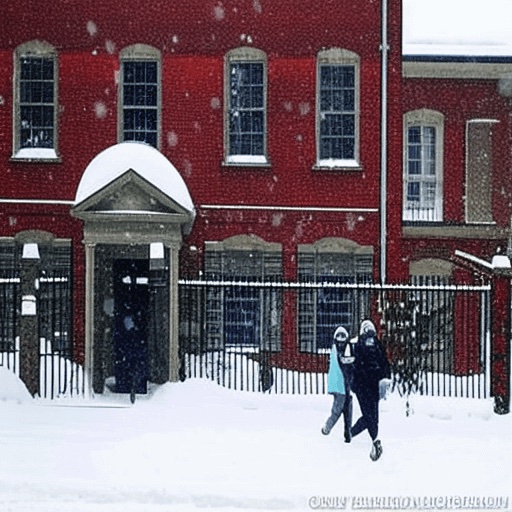}& 
\includegraphics[width=0.14\linewidth,height=0.08\linewidth]{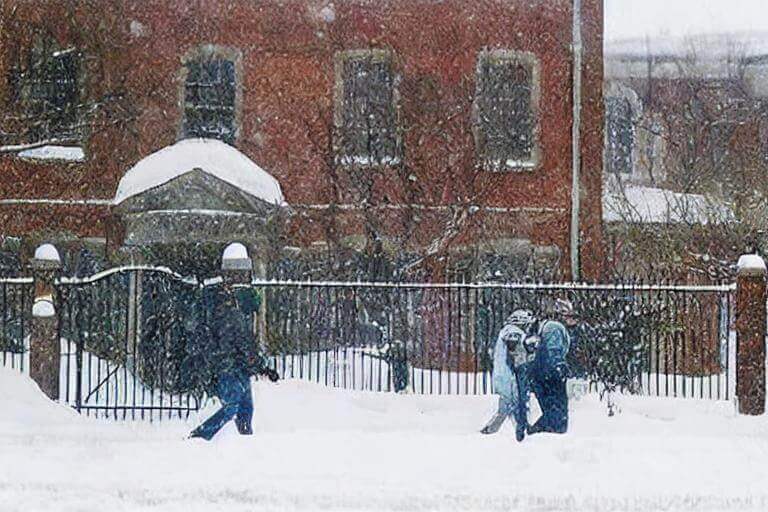}& 
\includegraphics[width=0.14\linewidth,height=0.08\linewidth]{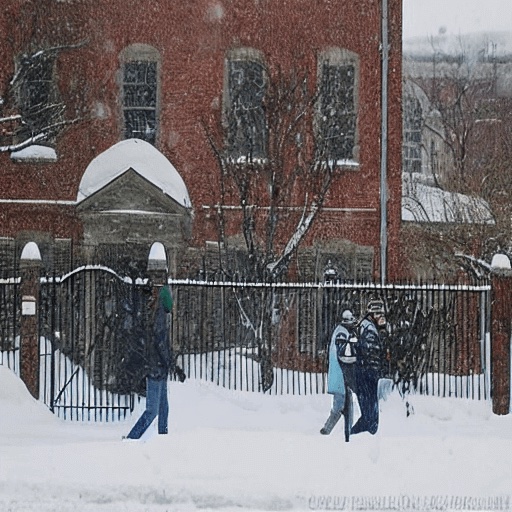}& 
\includegraphics[width=0.14\linewidth,height=0.08\linewidth]{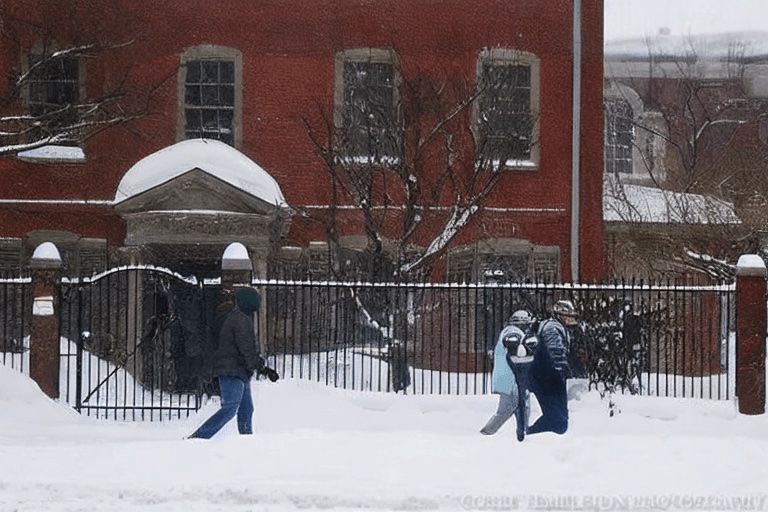}& 
\vspace{0.5mm}\\
\includegraphics[width=0.14\linewidth,height=0.08\linewidth]{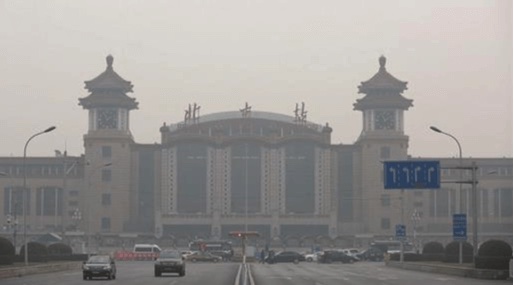}& 
\includegraphics[width=0.14\linewidth,height=0.08\linewidth]{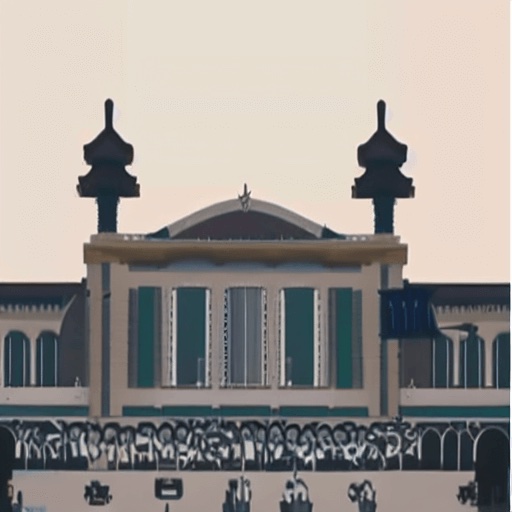}& 
\includegraphics[width=0.14\linewidth,height=0.08\linewidth]{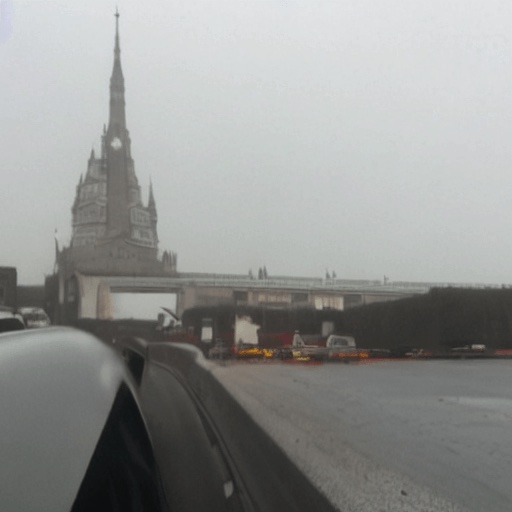}& 
\includegraphics[width=0.14\linewidth,height=0.08\linewidth]{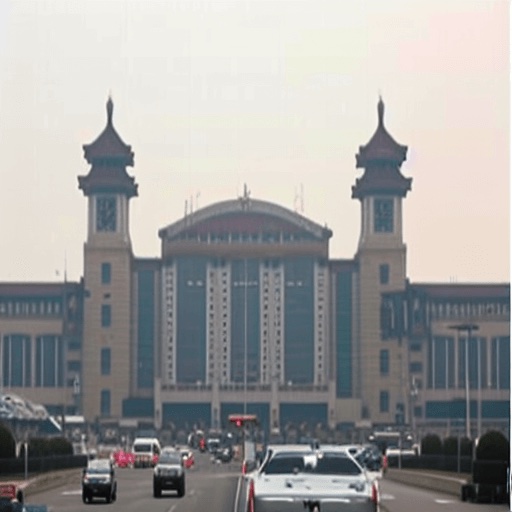}& 
\includegraphics[width=0.14\linewidth,height=0.08\linewidth]{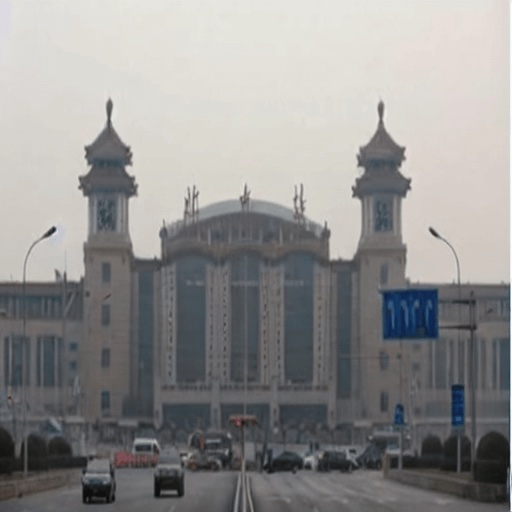}& 
\includegraphics[width=0.14\linewidth,height=0.08\linewidth]{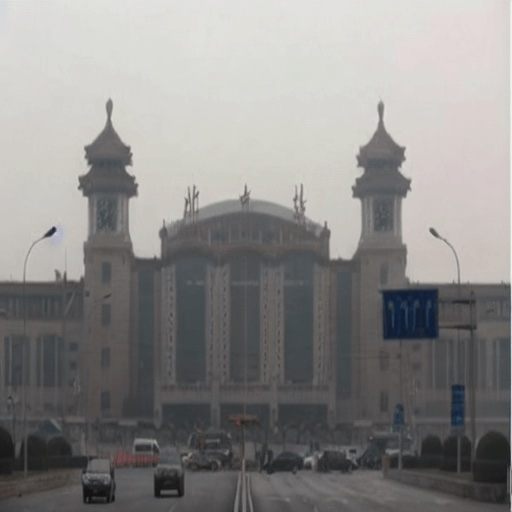}& 
\includegraphics[width=0.14\linewidth,height=0.08\linewidth]{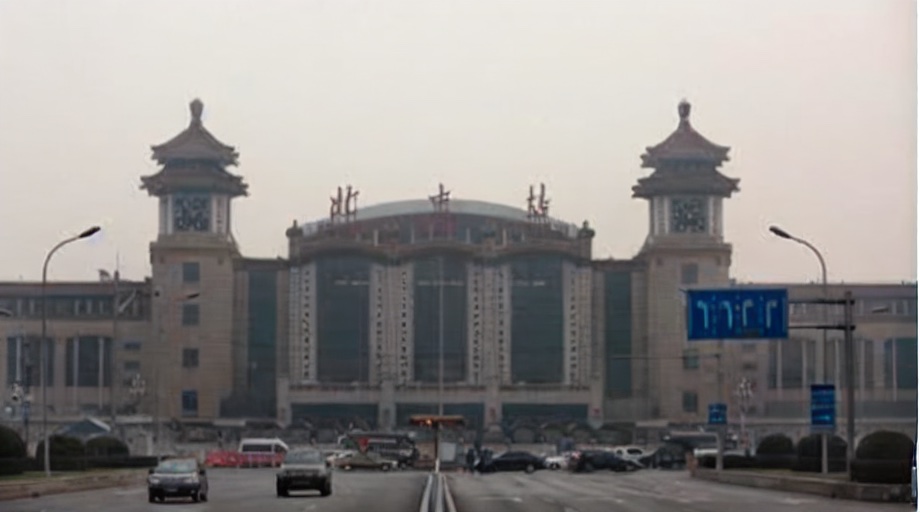}& 
\vspace{-0.5mm}\\
\fontsize{9pt}{\baselineskip}\selectfont  Input &
\fontsize{9pt}{\baselineskip}\selectfont  \ding{192} Inversion+Edit. &
\fontsize{9pt}{\baselineskip}\selectfont  \ding{193} TPB &
\fontsize{9pt}{\baselineskip}\selectfont  \ding{194} TPB+Inversion &
\fontsize{9pt}{\baselineskip}\selectfont \ding{195} SCB &
\fontsize{9pt}{\baselineskip}\selectfont \ding{196} TPB+SCB (Rec.) &
\fontsize{9pt}{\baselineskip}\selectfont  Ours \\

\end{tabular}
}
\end{center}
\vspace{-7mm}
\caption{Visual comparison of various Task-Plugin design variants. Row 1 and Row 2 showcase desnowing and dehazing, respectively.}
\label{fig:ablation_plugin}
\vspace{-3mm}
\end{figure*}

\subsection{Ablation Study}
\label{subsec:ablation}

\noindent \textbf{Task-Plugin.} 
We first evaluate the efficacy of Task-Plugins by exploring various ablated designs and comparing their performances on desnowing and dehazing. Unless specified otherwise, random noise is used during inference. \ryn{We have five ablated models.}
\ding{192} Inversion + Editing: DDIM Inversion with a task-specific description (\eg, ``\textit{a photo of a snowy day}") inverts the input image into an initial noise, retaining content. This is followed by editing using a target description (\eg, ``\textit{a photo of a sunny day}").
\ding{193} TPB: \fang{The SCB is removed,}
focusing solely on TPB training.
\ding{194} TPB + Inversion: Only TPB is trained, but DDIM Inversion is used for initial noise during inference.
\ding{195} SCB: The \fang{TPB} is removed to train \ryn{the SCB exclusively}.
\ding{196} TPB + SCB (Reconstruction): Training begins with SCB using self-reconstruction denoising loss, and then proceeds to TPB training \ryn{with the fixed SCB}. 
Performance results and comparison are presented in  \figref{fig:ablation_plugin} and \tableref{tab:ablation_plugin}.

We have the following observations. 
\ding{192} Inversion + Editing captures the global structure of the input image but loses detailed content.
\ding{193} TPB provides task-specific visual guidance but lacks spatial content constraints due to its focus on advanced features only. 
\ding{194} \yu{TPB, using inverted initial noise, excels in structured scenes (\eg, large buildings) but tends to deepen colors and create random content for smaller objects.} 
\ding{195} \yu{SCB maintains content details, but without task-specific visual guidance, it struggles to effectively remove degradations (\eg, snow or haze).}
\ding{196} TPB, when combined with reconstruction-based SCB, preserves image content through reconstruction while relying solely on TPB to address degradation. However, as SCB reintroduces all image features in each diffusion iteration, including original degradations (\eg, haze in row-2 of \figref{fig:ablation_plugin}), it inadvertently compromises the desired outcomes. 
Finally, incorporating the task-specific priors from both TPB and SCB in our Task-Plugin enables high-fidelity low-level task processing.

\begin{table}
\centering
\footnotesize
\setlength{\tabcolsep}{5.5pt}
\renewcommand\arraystretch{1.2}
\begin{tabular}{cl|ccc}
\shline
& Methods $\backslash$ FID $\downarrow$ &  Desnowing & Dehazing  \\
 \hline
\ding{192}  & Inversion + Editing & 48.54 & 35.05 \\  
\ding{193} & TPB  & 36.02 & 37.73 \\ 
\ding{194} &  TPB + Inversion  & 34.87 & 33.05 \\ 
\ding{195} &  SCB  & 34.71 &36.16  \\   
\ding{196} &   TPB + SCB (Reconstruction) & 34.50 &35.94  \\  
 &  TPB + SCB (Ours)  & 34.30 &34.68  \\ 
 \shline
\end{tabular}
\vspace{-3mm}
\caption{Ablation studies of variant Task-Plugin designs on two tasks: desnowing, dehazing. \yu{Note that although some variants \ryn{have} much lower FID scores, they tend to generate random content \fang{(refer to \ding{192}-\ding{194} of \figref{fig:ablation_plugin})}. 
In contrast, our final model guarantees both content fidelity and robust metric performances.}
}
\label{tab:ablation_plugin}
\end{table}

\begin{table}[t!]
    \centering
    \footnotesize
    \setlength{\tabcolsep}{3pt}
    \renewcommand\arraystretch{1.2}
    \begin{tabular}{l|cccc|cccc}
        \shline
        \multirow{2}{*}{Metrics} & \multicolumn{4}{c|}{Encoder}  & \multicolumn{4}{c}{Decoder} \\
          & E-1 & E-2 & E-3 & E-4 & D-4 & D-3 & D-2 & D-1 \\
        \hline
        FID  $\downarrow$ & 34.33 & 34.46  &  36.58   &  37.41  & 37.71 &34.59&34.20& 34.30       \\
        KID  $\downarrow$ & 5.23  & 5.52   &  7.18    &  7.84   & 7.57 & 5.55 &  5.20   &   5.20    \\
        \hline
        Param.(MB) & 14.88 &48.77&\multicolumn{4}{|c|}{182.31} &48.77 & 14.88 \\
        \shline
    \end{tabular}
    \vspace{-3mm}
    \caption{Ablation studies on the placement of SCB within the pre-trained SD's Encoder/Decoder stages on desnowing. `E/D-{$i$}' represents the $i$-th stage, with higher numbers indicating deeper layers. We modify the feature dimension in SCB to suit various stages of the pre-trained SD model, resulting in varied parameters.}
    \vspace{-2mm}
    \label{tab:plugin_position}
\end{table}

We also confirm the placement of SPB within the pre-trained SD model on desnowing task and show the results in Table \ref{tab:plugin_position}. 
Obviously, we can observe that for both the encoder and decoder of the pre-trained SD \cite{rombach2022high}, the fidelity diminishes and performance progressively decreases from the shallower to the deeper stages (\eg, stages 1 to 4). Thus, we inject the spatial features into the final stage of the decoder, balancing performance and parameters. \yu{Notably, the parameters of Task-Plugin module is only 1.67\% of the SD.}

\noindent \textbf{Plugin-Selector.}
As shown in \tableref{tab:selector}, we first evaluate the \ryn{accuracy of Plugin-Selector} in both single-task and multi-task scenarios (row-1 and -2), and observe consistently high accuracy. 
In addition, in a significantly extensive test with 120,000 samples (denoted as Multi-task*), it achieves an mAP accuracy of 0.936, demonstrating its effectiveness. Further, in a robustness test (denoted as Single + Non.) combining task-specific and task-irrelevant texts, it still achieves a notable zero-toleration accuracy of 0.779.

We also conduct an ablation study on the Plugin-Selector to evaluate the significance of each component, with results detailed in \tableref{tab:ablation_selector}. 
\ding{192} We remove the visual and textual projection heads separately. 
\ding{193} We assess the impact of varying the number of negative samples for contrastive training. 
The results first reveal that both visual and textual projection heads are crucial. Omitting the visual head results in training collapse and \textit{NaN} output, while removing the textual head \ryn{lowers the} ZTA \fang{metric} by 15.4\%. 
It also shows that increasing the number of negative samples (\eg, from $N=1$ to 15) consistently enhances selection accuracy\footnote{\noindent The default batch size is 8, implying 7 neg. samples \fang{and 1} pos. sample.}.

\begin{table}[t!]
    \centering
    \footnotesize
    \setlength{\tabcolsep}{3.5pt}
    \renewcommand\arraystretch{1.2}
    \begin{tabular}{l|cccccccc}
        \shline
         Tasks & ZTA $\uparrow$ &CP $\uparrow$& OP $\uparrow$ & OR  $\uparrow$ & CF1 $\uparrow$ & OF1 $\uparrow$ & mAP $\uparrow$ \\
        \hline
        Single-task & 0.998&-&0.998&-&-&0.998&0.998\\
        Multi-task    &0.979   &0.988 & 0.988 &0.927&0.956 &0.956&0.933     \\
        Multi-task* &0.969 & 0.983 &0.983 & 0.936 & 0.960&0.959&0.936    \\
        Single + Non. & 0.779  &0.814 & 0.808 &0.941&0.872 &0.870&0.775   \\
        \shline
    \end{tabular}
    \vspace{-3mm}
    \caption{Quantitative evaluation of the proposed Plugin-Selector. Asterisks (*) denotes more sample combinations. A dash (-) indicates \ryn{metric not applicable}. `Single + Non' refers to random combinations of single-task text inputs with non-existing (\ie, plugin-irrelevant) tasks, \ryn{to test} the Plugin-Selector's robustness.}
    \label{tab:selector}
\end{table}

\begin{table}[t!]
    \centering
    \footnotesize
    \setlength{\tabcolsep}{3.5pt}
    \renewcommand\arraystretch{1.3}
    \begin{tabular}{l|cc|ccccc}
        \shline
        \multirow{2}{*}{Single + Non.}&\multicolumn{2}{c|}{Remove} & \multicolumn{5}{c}{Number of \fang{Negative Samples}} \\
        & $\textit{VP}(\cdot)$& $\textit{TP}(\cdot)$ & 1 & 3 & 5 & 7 & 15\\
        \hline
        ZTA $\uparrow$ &\textit{NaN} &0.625 &0.559 & 0.648 & 0.725 & 0.779 & 0.817 \\
        
        \shline
    \end{tabular}
    \vspace{-3mm}
    \caption{Ablation studies of Plugin-Selector. `\textit{NaN}' indicates non-convergence of training, resulting in unavailable result. 
    }
    \label{tab:ablation_selector}
\end{table}

\noindent \textbf{Diverse Applications.} \figref{fig:application} demonstrates the versatility of \textit{Diff-Plugin}.  
Row-1 exemplifies complex, low-level task execution via sub-task integration (\eg, old photo restoration can be roughly divided into restoration and colorization.). 
 Row-2 highlights its ability to invert low-level tasks, enabling the generation of special effects like rain and snow.

\begin{figure}[t!]
\begin{center}
\begin{tabular}{c@{\hspace{0.5mm}}c@{\hspace{0.5mm}}c@{\hspace{0.5mm}}c@{\hspace{0.5mm}}c@{\hspace{0.5mm}}}
\includegraphics[width=0.20\linewidth,height=0.16\linewidth]{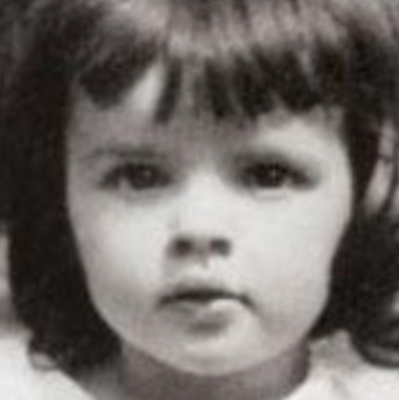}& 
\includegraphics[width=0.20\linewidth,height=0.16\linewidth]{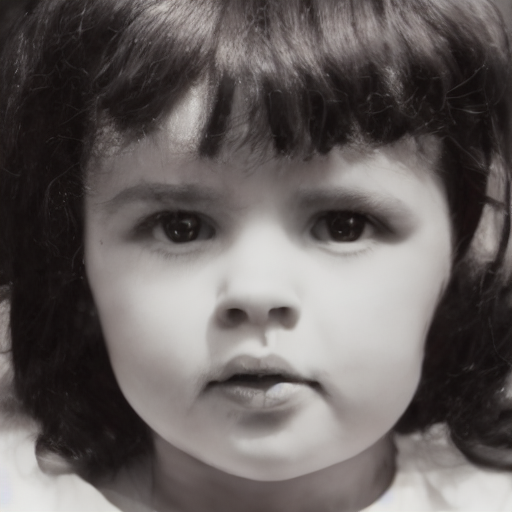}&
\includegraphics[width=0.20\linewidth,height=0.16\linewidth]{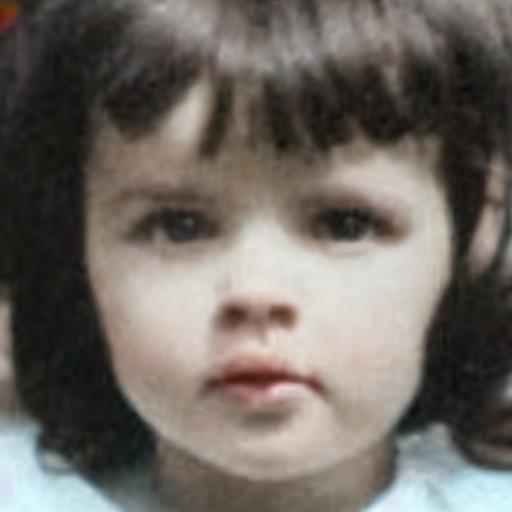}& 
\includegraphics[width=0.20\linewidth,height=0.16\linewidth]{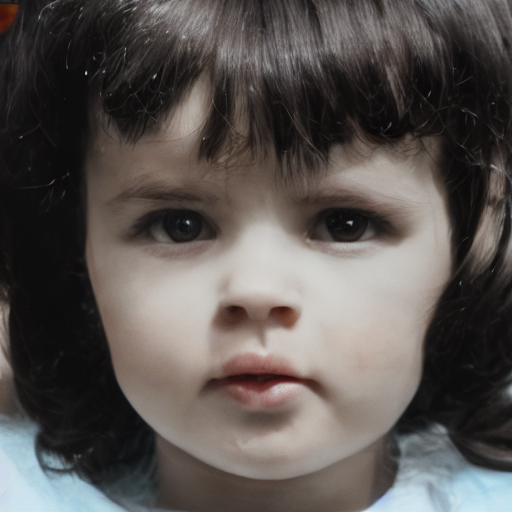}& 
\vspace{-1mm}\\
\fontsize{7pt}{\baselineskip}\selectfont  Input &
\fontsize{7pt}{\baselineskip}\selectfont  Restoration &
\fontsize{7pt}{\baselineskip}\selectfont  Colorization &
\fontsize{7pt}{\baselineskip}\selectfont  Restor. + Colori. \vspace{1mm}\\
\includegraphics[width=0.20\linewidth,height=0.16\linewidth]{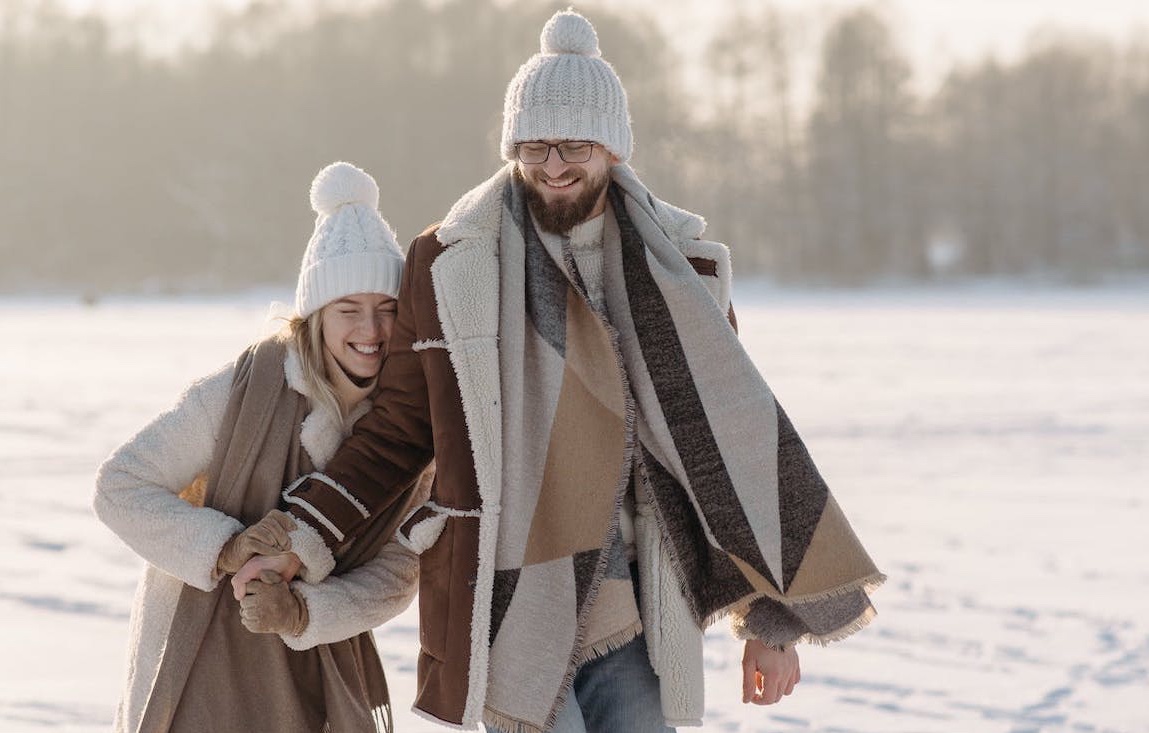}& 
\includegraphics[width=0.20\linewidth,height=0.16\linewidth]{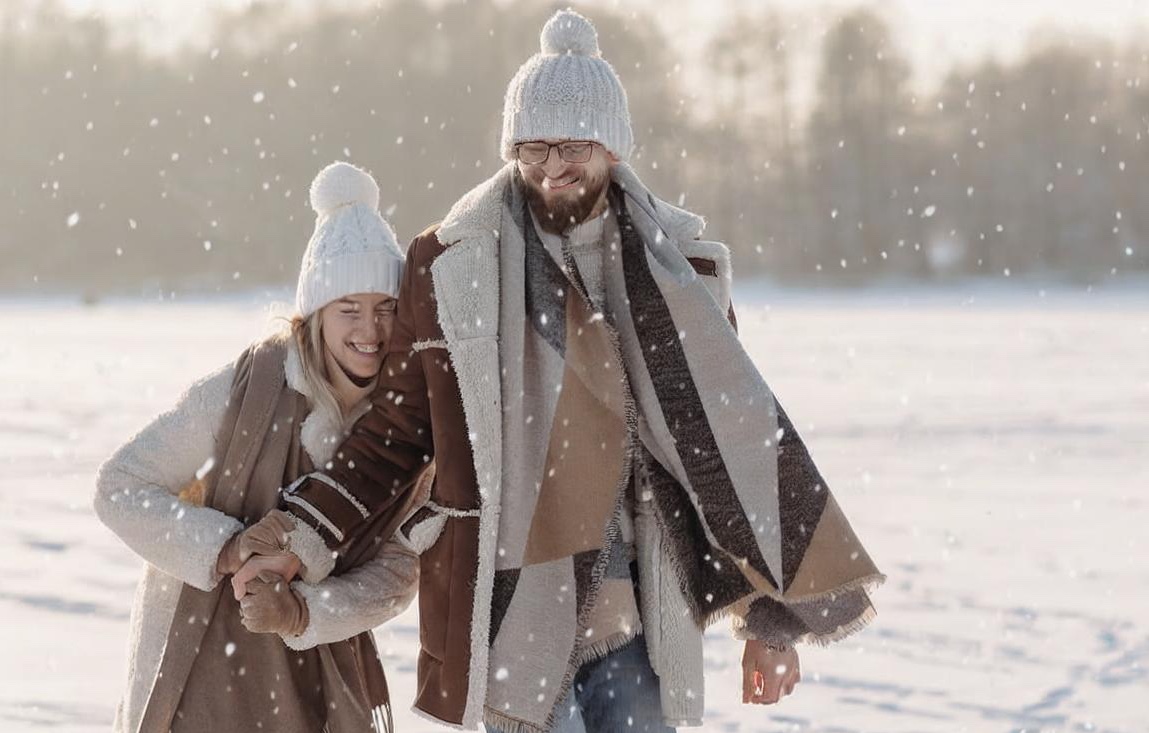}&
\includegraphics[width=0.20\linewidth,height=0.16\linewidth]{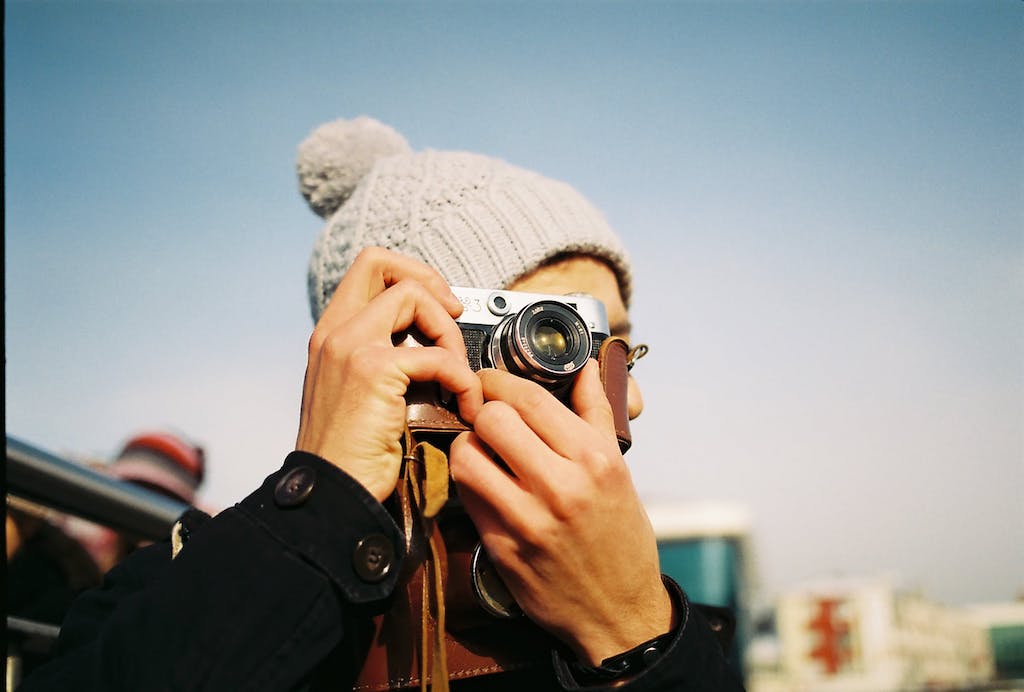}& 
\includegraphics[width=0.20\linewidth,height=0.16\linewidth]{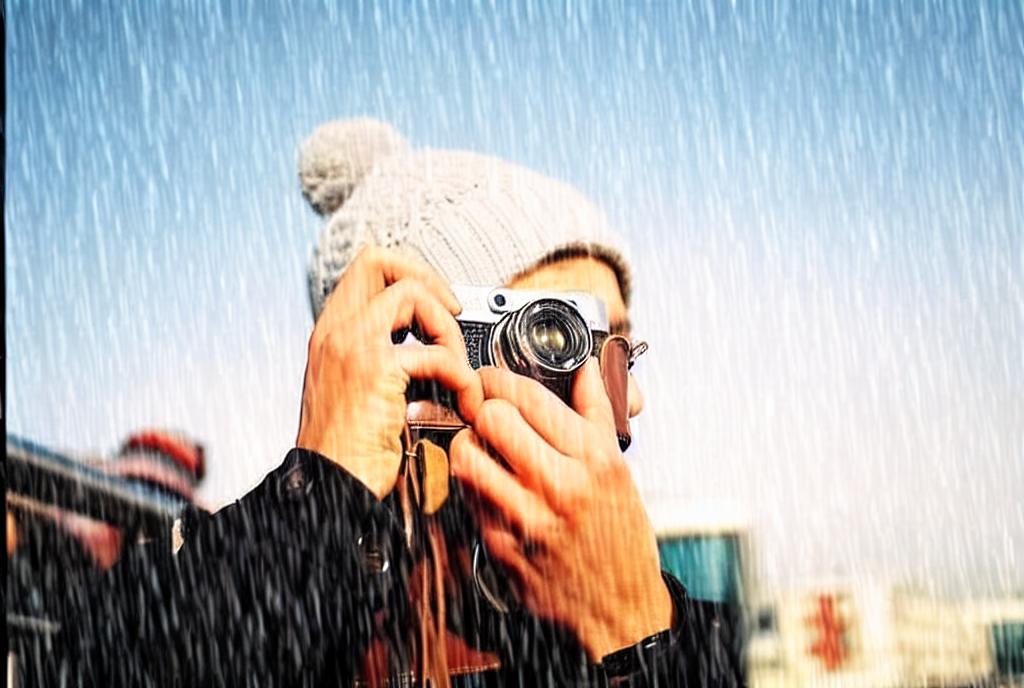}& 
\vspace{-1mm}\\
\fontsize{7pt}{\baselineskip}\selectfont  Input &
\fontsize{7pt}{\baselineskip}\selectfont  Snow Generation &
\fontsize{7pt}{\baselineskip}\selectfont  Input &
\fontsize{7pt}{\baselineskip}\selectfont  Rain Generation \\
\end{tabular}
\end{center}
\vspace{-7mm}
\caption{\ryn{Diverse} uses of \textit{Diff-Plugin}: multi-task combination in row-1 and reversed low-level tasks in row-2.} 
\vspace{-3mm}
\label{fig:application}
\end{figure}
\section{Conclusion}
\label{sec:cond}

In this paper, we presented \textit{Diff-Plugin}, a \ryn{novel} framework tailored for enhancing pre-trained diffusion models in \yu{handling various low-level vision tasks that need stringent details preservation.} Our Task-Plugin module, with its dual-branch design, effectively incorporates task-specific priors into the diffusion process to allow for high-fidelity details-preserving visual results without retraining the base model for each task. The Plugin-Selector further adds intuitive user interaction through text inputs, \ryn{enabling} text-driven low-level tasks and enhancing the framework's practicality. Extensive experiments across various vision tasks demonstrate \ryn{the superiority of our framework} over existing methods, especially in real-world scenarios.

One limitation of our current \textit{Diff-Plugin} framework is the inability in local editing. For example, in \figref{fig:teaser}, our method may fail to \ryn{remove only the snow specifically on the river while keeping} those in the sky. One possible solution for this problem is to integrate LLMs \cite{zhu2023minigpt,liu2023visual} to indicate the region in which the task is performed.

{
    \small
    \bibliographystyle{ieeenat_fullname}
    \bibliography{main}
}

\end{document}